\documentclass{article}

\usepackage{arxiv}

\usepackage[utf8]{inputenc} 
\usepackage[T1]{fontenc}    
\usepackage{hyperref}       
\usepackage{url}            
\usepackage{booktabs}       
\usepackage{amsfonts}       
\usepackage{nicefrac}       
\usepackage{microtype}      
\usepackage{lipsum}
\usepackage{graphicx}
\graphicspath{ {./images/} }

\usepackage[numbers]{natbib}
\usepackage{graphicx}
\usepackage{amsmath}
\usepackage{amssymb}
\usepackage{algorithm}
\usepackage{algorithmic}

\usepackage{times}  
\usepackage{helvet}  
\usepackage{courier}  

\usepackage{caption} 
\usepackage{appendix}
\graphicspath{ {./figures/} }
\usepackage{subfig}

\usepackage{newfloat}
\usepackage{listings}

\title{\textsc{Cradle-VAE}: Enhancing Single-Cell Gene Perturbation Modeling with Counterfactual Reasoning-based Artifact Disentanglement}

\author{
Seungheun Baek\thanks{Equal Contributors} \\
  Department of Computer Science\\
  Korea University\\
  Seoul, South Korea\\
  \texttt{sheunbaek@korea.ac.kr} \\
   \And
 Soyon Park\footnotemark[1] \\
  Department of Computer Science\\
  Korea University\\
  Seoul, South Korea\\
  \texttt{soyon\_park@korea.ac.kr} \\
  \And
 Yan Ting Chok \\
  Department of Computer Science\\
  Korea University\\
  Seoul, South Korea\\
  \texttt{yanting1412@korea.ac.kr} \\
  \And
 Junhyun Lee \\
  Department of Computer Science\\
  Korea University\\
  Seoul, South Korea\\
  \texttt{ljhyun33@korea.ac.kr} \\
  \And
 Jueon Park \\
  Department of Computer Science\\
  Korea University\\
  Seoul, South Korea\\
  \texttt{jueon\_park@korea.ac.kr} \\
  \And
 Mogan Gim\thanks{Corresponding Authors} \thanks{This work was done while the author was a postdoctoral researcher at Korea Univeristy} \\
  Department of Biomedical Engineering\\
  Hankuk University of Foreign Studies\\
  Yongin, South Korea\\
  \texttt{gimmogan@hufs.ac.kr} \\
  \And
 Jaewoo Kang\footnotemark[2] \\
  Department of Computer Science\\
  Korea University\\
  Seoul, South Korea\\
  \texttt{kangj@korea.ac.kr} \\
}

\begin{document}

\maketitle
\def\modelname{\textsc{Cradle-VAE}}

\begin{abstract}
Predicting cellular responses to various perturbations is a critical focus in drug discovery and personalized therapeutics, with deep learning models playing a significant role in this endeavor. Single-cell datasets contain technical artifacts that may hinder the predictability of such models, which poses quality control issues highly regarded in this area. To address this, we propose \modelname, a causal generative framework tailored for single-cell gene perturbation modeling, enhanced with counterfactual reasoning-based artifact disentanglement. Throughout training, \modelname~models the underlying latent distribution of technical artifacts and perturbation effects present in single-cell datasets. It employs counterfactual reasoning to effectively disentangle such artifacts by modulating the latent basal spaces and learns robust features for generating cellular response data with improved quality. Experimental results demonstrate that this approach improves not only treatment effect estimation performance but also generative quality as well. The \modelname~codebase is publicly available at https://github.com/dmis-lab/CRADLE-VAE.
\end{abstract}


\section{Introduction}
Understanding cellular responses to gene perturbations is crucial for identifying potential therapeutic targets. Single-cell technologies such as Perturb-seq~\citep{dixit2016perturb} have facilitated application of machine learning methodologies in addressing this task due to their high-resolution and high-throughput production of single-cell RNA sequencing (scRNA-seq) data. 

Previous works have proposed various computational methods for effectively modeling single-cell gene perturbation outcomes (i.e., treatment effects), mostly involving prediction of scRNA-seq gene expression profiles. One line of work features explicitly modeling the gene-gene relationships, incorporating prior knowledge graphs or networks inferred from the transcriptional data~\citep{roohani2024predicting, cui2024scgpt}. Another centralizes around employing variational autoencoders (VAE) which learn causal representations of single cells through modeling the disentanglement of its perturbation effects~\citep{lopez2023learning}. SAMS-VAE models the addition of two disentangled factors which are perturbation-independent cell representation (i.e., basal state) and sparse latent effects of gene perturbations (i.e., intervention)~\citep{bereket2024modelling}.

\begin{figure}[!t]
\centering
\includegraphics[width=0.8\columnwidth]{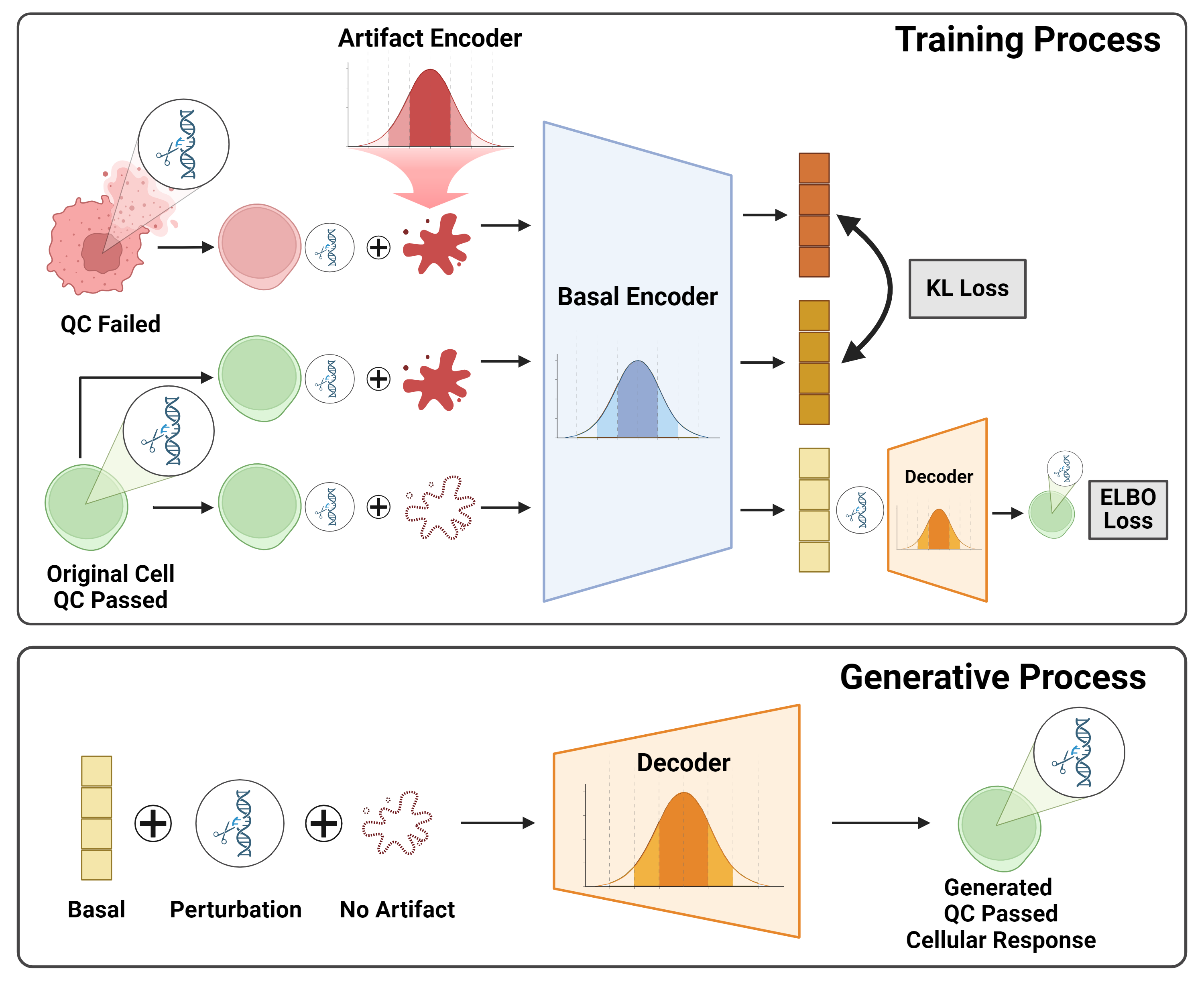}
\caption{Illustration of the training process and generative process of \modelname.}

\end{figure}

Despite the endeavor in improving performance in predicting cellular responses, the quality of training data used in previous works or data generated by their proposed models is not adequately evaluated. scRNA-seq datasets suffer from quality issues which are attributed to the limitations of existing sequencing protocols related to measurement of cells being stressed, broken, or killed. Some data might also correspond to empty droplets or droplets containing multiple cells (i.e., doublets)~\citep{ilicic2016classification}. Conventional quality control (QC) guidelines state that these data are deemed \textit{under-qualified} and the distortions that arise from the limitations of the scRNA-seq protocols are said to be \textit{technical artifacts}~\citep{hong2022comprehensive}. 

A straightforward way to tackle data quality issues caused by the technical artifacts would be resorting to filtering scRNA-seq data based on QC criteria. This method involves excluding QC failed data that may confound downstream analyses and interpretation~\citep{analysisguide2022tenex}. In fact, both the quantity and the quality of scRNA-seq data, from which the model learns the data distribution, strongly influences that of the model performance~\citep{chen2023softmatch}. This implies a trade-off between the strictness of gene expression data quality control and the abundance of training data required for effective generalization~\citep{heumos2023best}.

Inspired by recent efforts in disentangling the latent gene perturbation effects from the given scRNA-seq data via the VAE framework~\citep{lopez2023learning}, we propose a similar approach for handling its inherent artifacts as well. Instead of removing the QC failed data samples, we can implement a module that disentangles the inherent technical artifacts from those samples, which ultimately leads to better generative quality while preserving the limited number of scRNA-seq gene expression profiles in the training dataset. This deeply relates to counterfactual reasoning, as our proposed approach not only answers the question \textit{what will the generative outcome be if given this gene perturbation instead?} but also \textbf{under this specific gene perturbation, what would the generative outcome have been if technical artifacts had been absent?}

In this work, we propose \modelname, a novel VAE framework designed to learn causal representations of scRNA-seq data by utilizing \textbf{C}ounterfactual \textbf{R}easoning-based \textbf{A}rtifact \textbf{D}isentang\textbf{LE}ment. \textsc{Cradle-VAE} aims to address quality issues of both training and generated data by disentangling technical artifacts from the natural, perturbation-independent variation in cells through counterfactual reasoning. Specifically, given a QC passed scRNA-seq gene expression profile (i.e., artifact-free) as input, \modelname~uses an auxiliary loss objective that guides the encoded counterfactual basal state (i.e., artifact-present) towards its reference counterfactual basal state. The latter is constructed as an aggregation of QC failed scRNA-seq data samples under the same gene perturbation treatment.

Our experiments demonstrate that compared with its baselines and ablations, \modelname~generates gene expression profiles deemed as cellular response predictions that not only showcase superior correlation but also generative quality measured by QC pass rate. To the best of our knowledge, it is the first attempt to model the presence of technical artifacts in scRNA-seq datasets for perturbation response prediction and exploit them leveraging counterfactual reasoning to improve generative quality. The main contributions of this work are summarized as follows:
\begin{itemize}
\item We propose \modelname, a novel VAE-based cellular response prediction model that addresses quality issues in the realm of scRNA-seq data.
\item We introduce an auxiliary loss objective that guides \modelname's disentanglement of artifacts during the training process.
\item Experimental results show that \modelname~robustly predicts cellular responses by generating gene expression profiles with higher quality compared to previous methods especially when given unseen perturbations as input.
\item Qualitative analysis highlights how our proposed approach contributes to enhancing \modelname's disentanglement ability improving its generative quality.
\end{itemize}

\section{Related Works}
\subsection{Disentanglement in Single-cell Perturbation Response Prediction}
Recent advancements in single-cell RNA sequencing technologies have significantly enhanced our understanding of cellular responses to chemical and genetic perturbations~\citep{srivatsan2020massively, norman2019exploring}. Due to the complexity of studying the phenotypic effects of cellular perturbations and their underlying factors, previous works have focused on leveraging causal learning which aims to understand the mechanisms by which variables influence each other and predicting the outcome of interventions~\citep{spirtes2010introduction}. CPA utilizes a disentanglement strategy based on adversarial approach~\citep{lotfollahi2023predicting}. Moreover, with VAEs being the primary generative models, studies have focused on disentangling the latent variables that constitute the true distribution of scRNA-seq data. Both sVAE+~\citep{lopez2023learning} and SAMS-VAE~\citep{bereket2024modelling} utilize sparse mechanism shifts to disentangle gene perturbations.
\subsection{Counterfactual Reasoning in Single-cell Perturbation Response Prediction}
Another line of previous work focuses on employing counterfactual reasoning in predicting the outcomes of single-cell gene perturbations. Counterfactual reasoning helps generative models such as VAEs expand their understanding in causal relationships between different factors such as gene-gene interactions. GraphVCI adopted this concept in enhancing the individuality of cellular responses and dynamically modulating the graph regulatory network structure based on different gene perturbations~\citep{wu2022predicting}. Similarly, CODEX incorporates the counterfactual reasoning approach in predicting the genetically perturbed scRNA-seq data given the unperturbed data (i.e., control expression profile) along with dosage information and specific interventions as input. \\

None of the previous models have explicitly considered data quality issues caused by scRNA-seq protocols despite being emphasized in biology domain. Our study addresses this by incorporating counterfactual reasoning related to the presence of latent technical artifacts in scRNA-seq data so that the generative model effectively disentangles them during its training process. 

\section{Methods}
\subsection{scRNA-seq Dataset}
We define a $N$-sized scRNA-seq dataset $(x_i, p_i, a_i)_{i=1}^N$ where each data instance includes a gene expression vector $x_i\in\mathbb{R}^{D_x}$, a gene perturbation vector $p_i\in\left\{0,1\right\}^{T}$ and an artifact presence label $a_i\in\left\{0,1\right\}$ where $D_x$ is the total number of genes used in this task, and $T$ is the number of perturbation types. Each bit in $p_i$ specifies whether its corresponding gene was perturbed prior to obtaining $x_i$. Also, $a_i$ indicates the presence of technical artifacts in $x_i$. In our task's context, $x_i$ is the cellular response when given treatment $p_i$. If $x_i$ passes a predefined quality control criteria, then $a_i=0$ ; otherwise, $a_i=1$. 

\subsection{Quality Control Criteria}
We elaborate the process of labeling each expression vector with $a_i$ based on our established quality control (QC) criteria. Having adopted the filtering guidelines provided by Scanpy and 10X Genomics, we established the following six QC sub-criteria: UMI counts, number of features, percent of mitochondrial (mt) reads, percent of hemoglobin reads (hb), percent of ribosomal (rb) reads and doublet detection~\citep{wolf2018scanpy,analysisguide2022tenex}. The first five sub-criteria are determined using data-driven thresholds calculated as scaled median absolute deviation (MAD)~\cite{ocasio2019scrna, you2021benchmarking} while the last criterion is a binary label identified by Scrublet~\citep{wolock2019scrublet}. We used three to five times of the MAD ($3\sigma$, $4\sigma$, $5\sigma$) since threshold selection can vary across studies~\citep{ocasio2019scrna, you2021benchmarking}, where $3\sigma$ represents the strictest QC cut-off, followed by $4\sigma$ and $5\sigma$.
\subsection{\modelname}

\subsubsection{Encoder Module} 
\begin{figure*}
\begin{minipage}{\textwidth}
    \centering
    \includegraphics[width=0.875\textwidth]{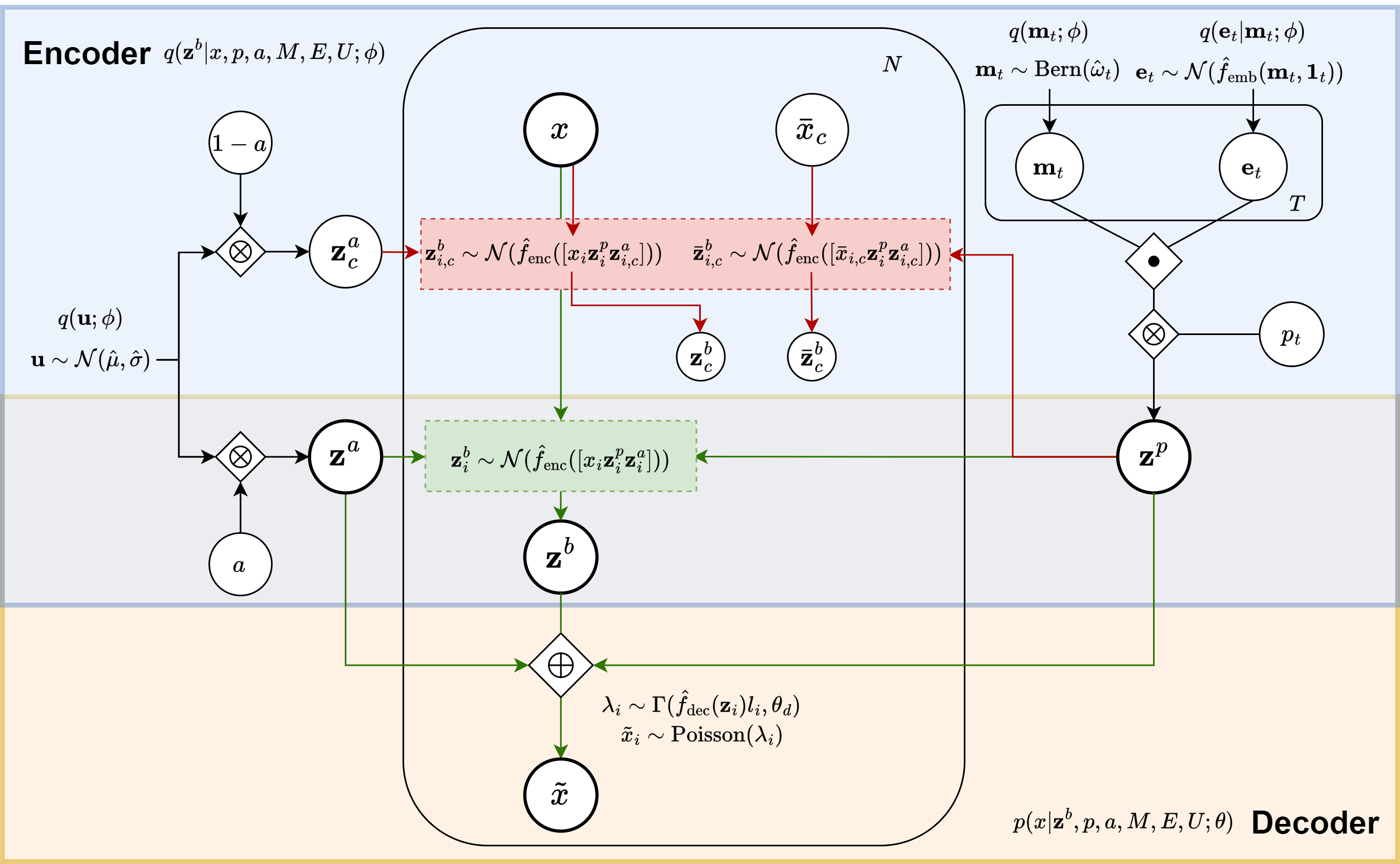}
\end{minipage}
\caption{Graphical model of \modelname. $\bullet$ represents Hadamard product operation; $\otimes$ represents matrix multiplication operation; $\oplus$ represents vector concatenation.}
\label{figure:cradlevae}
\end{figure*}

The overall architecture of \modelname~is shown in Figure~\ref{figure:cradlevae}. During training, the encoder part of \modelname~takes data instance $(x_i, p_i, a_i)$ as input and encodes it into three different latent representations which are latent basal state embedding $\mathbf{z}_i^b\in\mathbb{R}^{D_z}$, latent perturbation effect embedding $\mathbf{z}_i^p\in\mathbb{R}^{D_z}$ and latent artifact embedding $\mathbf{z}_i^a\in\mathbb{R}^{D_z}$ where $D_z$ is the dimension size of latent subspaces. The objective of this module is to disentangle these three latent variables and learn their individual contributions to the observed true data distribution. 

Algorithm~\ref{alg:encoding} shows \modelname's encoding process which inherits the formulation basis from \citeauthor{bereket2024modelling}'s work. The latent perturbation effect embedding $\mathbf{z}_i^p$ is an additive composition of global gene-wise perturbation effects, $\mathbf{e}_t$, induced by global sparse latent offsets, $\mathbf{m}_t$, which are sampled from parameterized prior Normal distribution and Bernoulli distribution, respectively (Algorithm~\ref{alg:encoding}.2,~3,~7). Similarly, the latent artifact embedding $\mathbf{z}_i^a$ is a multiplication of $a_i$ and $\mathbf{u}$, which is sampled from its own parameterized prior distribution (Algorithm~\ref{alg:encoding}.5,~8). 

$\mathbf{z}_i^b$ is sampled from a Normal distribution that is parameterized by a neural network $\hat{f}_{enc}$ taking $x_i$, $\mathbf{z}_i^p$ and $\mathbf{z}_i^a$ as input (Algorithm~\ref{alg:encoding}.12). $\mathbf{1}_t$ is the one-hot encoding of the $t$th gene perturbation treatment while both $\hat{f}_{emb}$ and $\hat{f}_{enc}$ are trainable neural networks.

\subsubsection{Decoder Module}
During training, the decoder part of \modelname~takes the latent embeddings ($\mathbf{z}_i^b,\mathbf{z}_i^p,\mathbf{z}_i^a$) as input and samples $\tilde{x}_i$ from a parameterized Gamma-Poisson distribution. Algorithm~\ref{alg:decoding} shows \modelname's decoding process where $\hat{f}_{dec}$ is a learnable neural network with final softmax layer that outputs the expected frequency for each gene used for parameterizing the Gamma-Poisson distribution. $l_i$ and ${\theta}_d$ denote the total number of read counts for the $i$th cell and learnable inverse dispersion used universally across all cells respectively.

\begin{figure}[!t]
\begin{minipage}[t]{0.5\columnwidth}
\begin{algorithm}[H]
\hsize=\textwidth
\caption{\modelname~Encoding Process}
\label{alg:encoding}
\begin{algorithmic}[1]
\REQUIRE{$X \in \mathbb{R}^{N \times D_x}$, $\bar{X}_{c} \in \mathbb{R}^{N \times D_x}$, $P \in \left\{0,1\right\}^{N \times T}$, $A \in \left\{0,1\right\}^{N}$}
\FOR {$t$ from $1$ to $T$}
\STATE $\mathbf{m}_t \sim \text{Bernoulli}(\mathbf{\hat{\omega}}_t)$
\STATE $\mathbf{e}_t \sim \mathcal{N}(\hat{f}_\text{emb}(\mathbf{m}_t,\mathbf{1}_t))$
\ENDFOR
\STATE $\mathbf{u} \sim \mathcal{N}(\mathbf{\hat{\mu}},\mathbf{\hat{\sigma}})$
\FOR{$i$ from $1$ to $N$}
\STATE $\mathbf{z}_i^p = \sum_{t=1}^T p_{i,t}(\mathbf{e}_t \odot \mathbf{m}_t)$
\STATE $\mathbf{z}_i^a = a_{i}\mathbf{u}$\\

\STATE $\mathbf{z}_{i,c}^a = (1-a_i)\mathbf{u}$
\STATE $\mathbf{z}_{i,c}^b \sim \mathcal{N} (\hat{f}_\text{enc}([x_i \oplus \mathbf{z}_i^p \oplus \mathbf{z}_{i,c}^a]))$
\STATE ${\mathbf{\bar{z}}_{i,c}^b} \sim \mathcal{N}(\hat{f}_\text{enc}([\bar{x}_{i,c} \oplus \mathbf{z}_i^p \oplus \mathbf{z}_{i,c}^a]))$
\STATE $\mathbf{z}_i^b \sim \mathcal{N}(\hat{f}_\text{enc}([x_i \oplus \mathbf{z}_i^p \oplus \mathbf{z}_i^a]))$

\ENDFOR
\end{algorithmic}
\end{algorithm}
\begin{algorithm}[H]

\caption{\modelname~Decoding Process}
\label{alg:decoding}
\begin{algorithmic}[1]
\REQUIRE $\mathbf{z}^b \in  \mathbb R^{N \times D_z}$, $\mathbf{z}^p\in\mathbb{R}^{N\times D_z}$, $\mathbf{z}^a\in\mathbb{R}^{N\times D_z}$
\FOR{$i$ from $1$ to $N$}
    \STATE $\mathbf{z}_i = [\mathbf{z}_i^b \oplus \mathbf{z}_i^p \oplus \mathbf{z}_i^a]$
    \STATE $\mathbf{\lambda}_i \sim \Gamma(\hat{f}_\text{dec}({\mathbf{z}_i}) l_i, {\theta}_d)$
    \STATE $\tilde{x}_i \sim \text{Poisson}(\mathbf{\lambda}_i)$
\ENDFOR
\end{algorithmic}
\end{algorithm}
\end{minipage}
\hspace{0.8cm}
\begin{minipage}[t]{0.5\columnwidth}

\begin{algorithm}[H]
\caption{\modelname~Generative Process}
\label{alg:generative}
\begin{algorithmic}[1]
\REQUIRE $P \in \left\{0,1\right\}^{N \times T}$
\FOR {$t$ from $1$ to $T$}
\STATE $\mathbf{m}_t \sim \text{Bernoulli}(\mathbf{\hat{\omega}}_t)$
\STATE $\mathbf{e}_t \sim \mathcal{N}(\hat{f}_\text{emb}(\mathbf{m}_t,\mathbf{1}_t))$
\ENDFOR
\STATE $\mathbf{u} \sim \mathcal{N}(\mathbf{\hat{\mu}},\mathbf{\hat{\sigma}})$
\STATE $\mathbf{z}_i^a = 0\mathbf{u}$
\FOR{$i$ from $1$ to $N$}
    \STATE $\mathbf{z}^b_i \sim \mathcal N (0,I)$
    \STATE $\mathbf{z}_i^p=\sum_{t=1}^T \mathbf{p}_{i,t}(\mathbf{e}_t\odot\mathbf{m}_t)$
    \STATE $\mathbf{z}_i = [\mathbf{z}_i^b \oplus \mathbf{z}_i^p \oplus \mathbf{z}_i^a]$
    \STATE $\mathbf{\lambda}_i \sim \Gamma(\hat{f}_\text{dec}(\mathbf{z}_i) l_i, {\theta}_d)$
    \STATE $\tilde{x}_i \sim \text{Poisson}(\mathbf{\lambda}_i)$
\ENDFOR
\end{algorithmic}
\label{algorithm:generation}
\end{algorithm}
\end{minipage}
\end{figure}

\newpage
\subsubsection{Variational Inference}
Considering the intractability of the data marginal probability $p(X|P,A)$, we define the correlated variational distribution $q(Z|X,P,A)$ by approximating the posterior distribution of latent variables:

\begin{align}
\begin{split}
    &q(Z^b,M,E,U|X,P,A) = \left(\prod_{t=1}^{T}q(\mathbf{e}_t|\mathbf{m}_t;\phi)q(\mathbf{m}_t;\phi) \right) \\
    &\quad\quad \times q(\mathbf{u};\phi)\left(\prod_{i=1}^{N} q(\mathbf{z}^b_i|x_i,p_i,a_i,M,E,U;\phi)\right)
\end{split}
\end{align}

for latent basal state embeddings $Z^b\in\mathbb{R}^{N \times D_z}$,  global latent perturbation masks $M\in\left\{0,1\right\}^{T \times D_z}$, global latent perturbation embeddings $E\in\mathbb{R}^{T \times D_z}$, global latent artifact embeddings $U\in\mathbb{R}^{1 \times D_z}$, gene expression matrix $X\in\mathbb{R}^{N \times D_x}$, gene perturbation matrix $P\in\left\{0,1\right\}^{N \times T}$, and artifact presence labels $A\in\left\{0,1\right\}^{N}$.

We employ stochastic variational inference~\citep{hoffman2013stochastic} to approximate the posterior distribution $\log p(X|P,A)$. The learnable parameters ($\theta$,$\phi$) of \modelname~are optimized by maximizing the evidence lower bound (ELBO) which is mathematically expressed as below:

\begin{align}
\begin{split}
&\mathcal{J}_{1}(\theta,\phi) = \mathbb{E}_{Z^b,M,E,U \sim q(\cdot|X,P,A;\phi)} \\
&\quad\quad\quad\quad\left[ \log \frac{p(X,Z^b,M,E,U|P,A;\theta)}{q(Z^b,M,E,U|X,P,A;\phi)} \right]
\end{split}
\end{align}
\vspace{0.1cm}
\subsubsection{Artifact Disentanglement by Counterfactual Reasoning}
We propose to exploit the counterfactual outcome of the same gene perturbation treatment as means to reinforce disentanglement of latent variables related to quality degradation caused by technical artifacts. We add the following modifications to \modelname's encoding process $x_i$ is a QC passed gene expression profile (i.e., $a_i=0$). 

First, \modelname~additionally builds a \textit{counterfactual latent artifact embedding} $\mathbf{z}^{a}_{i,c}=(1-a_i)\mathbf{u}$ which is opposite to $\mathbf{z}^{a}_{i}=a_i\mathbf{u}$ being zero-scaled (Algorithm \ref{alg:encoding}.9). It is then used for sampling the \textit{counterfactual latent basal state embedding} $\mathbf{z}^{b}_{i,c}$ from a Normal distribution parameterized by $\hat{f}_{enc}$ (Algorithm \ref{alg:encoding}.10). Meanwhile, for each QC passed gene expression profile $x_i$, we first sample its \textit{counterfactuals} from our dataset that share the same gene perturbation treatment but are QC failed. We then compute their median $\bar{x}_{i,c}$ to feed it along with $\mathbf{z}_i^p$ and $\mathbf{z}_{i,c}^a$ into the neural network $\hat{f}_{enc}$, from where we sample the \textit{reference counterfactual latent basal state embedding} $\mathbf{\bar{z}}_{i,c}^b$ (Algorithm \ref{alg:encoding}.11).

We imposed an auxiliary loss objective that guides $\mathbf{z}^{b}_{i,c}$ to be aligned with $\mathbf{\bar{z}}^{b}_{i,c}$. This is done by minimizing the Kullback–Leibler (KL) divergence between the two latent basal state embeddings which is mathematically expressed as follows:
\begin{align}
    \mathcal{J}_{2}(\phi) = -\text{KL}\left[ q(Z^b_c | X, P, A; \phi) \| q(\bar{Z}^b_c | \bar{X}, P, A; \phi) \right]
\end{align} \\

We expect the loss objective to provide two benefits for \modelname. First, the computed gradients that are back-propagated through $\hat{f}_{enc}$ to $\mathcal{N}(\hat{\mu},\hat{\sigma})$ exhibit additional supervision to the disentanglement of artifact-related latent variables, facilitating a clearer distinction between QC passed and QC failed cases. Second, the latent basal state embeddings that are encoded by $\hat{f}_{enc}$ help guide the $\hat{f}_{dec}$ to generate the data samples that not only correlate with the true cellular responses but are also more likely to pass the QC criteria. We will explore these benefits later through our quantitative experiments and qualitative analysis.

The overall learning objective that optimizes the trainable parameters $\theta$, $\phi$ is then defined as follows:
\begin{align}
    \mathcal{J}(\theta,\phi) = \mathcal{J}_{1}(\theta,\phi) + \alpha\mathcal{J}_{2}(\phi)
\end{align}
where $\alpha$ is the hyperparameter for controlling the alignment intensity of the auxiliary loss objective.

\subsubsection{Generative Process}
After training, \modelname~generates its predicted cellular responses by sampling the latent basal state embedding $\mathbf{z}_i^b$ from a normal distribution ($\mathcal{N}(0,I)$) and combining it with $\mathbf{z}_i^p$ and $\mathbf{z}_i^a$ sampled from the encoder module's parameterized distributions. Finally, $[\mathbf{z}_i^b\oplus\mathbf{z}_i^p\oplus\mathbf{z}_i^a]$ is fed to $\hat{f}_{dec}$, which generates the read counts for each gene (Algorithm~\ref{alg:generative}.11,12). Note that the global latent artifact embedding $\mathbf{u}$ is multiplied by $a_i=0$ since \modelname~is used to generate artifact-free gene expression data which is expected to pass the QC criteria (Algorithm~\ref{alg:generative}.6).

Formally, we define the joint probability distribution over the observed and latent variables as:

\begin{equation}
\begin{split}
    &p(X,Z^b,M,E,U|P,A;\theta) = \left(\prod_{t=1}^{T} p(\mathbf{m}_t)p(\mathbf{e}_t)\right)p(\mathbf{u})   \\
    &\quad\quad \times \left(\prod_{i=1}^{N} p(\mathbf{z}^b_i) p(x_i|\mathbf{z}^b_i,p_i,a_i,M,E,U; \theta)\right)
\end{split}
\end{equation}

\section{Experiments}

\subsection{Experiment Settings} 
We evaluated \modelname~on four Perturb-seq datasets, i.e. Norman dataset~\citep{norman2019exploring}, Dixit dataset~\citep{dixit2016perturb}, Replogle dataset~\citep{replogle2022mapping}, and Adamson dataset~\citep{adamson2016multiplexed}. We adopted the preprocessing approaches done to Replogle dataset from \citeauthor{lopez2023learning} and other datasets from \citeauthor{ji2021machine}. The details of each dataset are shown in Table~\ref{tab:datasets}.

\begin{table}[hbt!]
{\large
\resizebox*{0.5\columnwidth}{!}{%
\renewcommand{\arraystretch}{1.1}
\begin{tabular}{ccccc}
\hline
\textbf{Dataset}            & \textbf{\# of Cells} & \textbf{\# of Genes} & \textbf{\begin{tabular}[c]{@{}c@{}}\# of Perts\end{tabular}} & \textbf{Perturbation} \\ \hline
\textbf{Norman}        & 111,255        & 19,018         & 105 + \underline{131}                                                              & CRISPRa               \\
\textbf{Dixit}          & 103,420        & 18,531         & 10 + \underline{45}                                                                 & CRISPR-Cas9           \\ \hline
\textbf{Replogle} & 118,641        & 1,187          & 722                                                                    & CRISPRi               \\
\textbf{Adamson}       & 62,623         & 17,115         & 90                                                                     & CRISPRi           \\ \hline
\end{tabular}
}
\caption{Summary of Perturb-seq datasets used in our experiments. Notably, \citeauthor{norman2019exploring} and \citeauthor{dixit2016perturb} include multi-gene perturbations which is underlined, while \citeauthor{replogle2022mapping} and \citeauthor{adamson2016multiplexed} consist of only single-gene perturbations.}
\label{tab:datasets}
}
\end{table}

We compared \modelname~against four other causal learning-based VAE models, namely sVAE+~\citep{lopez2023learning}, CPA-VAE~\citep{bereket2024modelling}, SAMS-VAE~\citep{bereket2024modelling}, and conditional-VAE~\citep{sohn2015learning}. We additionally considered the variants of \modelname~trained under different QC threshold settings (3$\sigma$,4$\sigma$,5$\sigma$). Note that we applied the same QC criteria to all data instances partitioned into train, valid and testing purposes. 

In our evaluation, we considered the characteristics of data perturbations during the assessment process. For datasets involving multi-gene perturbations, the test set was constructed using combinations not encountered during training, representing approximately 25\% of the total possible combinations. Conversely, for datasets involving single perturbations, the evaluation emphasized the models' ability to capture trends in the observed data within the context of single-perturbation scenarios.

To robustly evaluate the models with respect to varying data quality, we trained and evaluated all baseline models with five different random seeds and reported their averaged results. Our main evaluation metric is the Average Treatment Effect Pearson Correlation (ATE-$\rho$) introduced by ~\citeauthor{bereket2024modelling}, that measures the correlation between model-predicted expression values and the experimental data across all genes. We also calculated the R-square score for the estimated average treatment effects as well (ATE-$R^2$). In addition, we employed the Jaccard similarity between top 50 model-predicted differentially expressed genes and true differentially expressed genes as defined in previous works~\citep{roohani2024predicting}.

As our work highlights the importance of addressing quality issues in scRNA-seq data, we formulated an evaluation metric that measures the model's generative quality, denoted as QC Pass Rate. The QC Pass Rate (QCPR) is calculated by dividing the number of generated data samples that passed the QC criteria divided by total number of generated data samples. Note that the threshold in QC criteria is equally applied for the annotation of Perturb-seq dataset and in the QCPR metric. 

\begin{table*}[hbt!]
{\large
\resizebox*{\textwidth}{!}{%
\renewcommand{\arraystretch}{1.15}

\begin{tabular}{cl|cccc|llll}
\hline
\multicolumn{2}{c|}{\textbf{Dataset}}                                                                                         & \multicolumn{4}{c|}{\textbf{Norman}}                                                                                                                                                                           & \multicolumn{4}{c}{\textbf{Dixit}}                                                                                                                                                                                                                     \\ \hline
\multicolumn{1}{c|}{Model}                     & \multicolumn{1}{c|}{\begin{tabular}[c]{@{}c@{}}QC threshold\end{tabular}} & \begin{tabular}[c]{@{}c@{}}ATE-$\rho$\end{tabular} & ATE-$R^2$                                       & Jaccard                                      & \begin{tabular}[c]{@{}c@{}}QCPR (\%)\end{tabular} & \multicolumn{1}{c}{\begin{tabular}[c]{@{}c@{}}ATE-$\rho$\end{tabular}} & \multicolumn{1}{c}{ATE-$R^2$}                   & \multicolumn{1}{c}{Jaccard}                  & \multicolumn{1}{c}{\begin{tabular}[c]{@{}c@{}}QCPR (\%)\end{tabular}} \\ \hline
\multicolumn{1}{l|}{Conditional VAE}           &                                                                              & {0.5314\scriptsize ± 0.04\normalsize}             & {\underline {0.2766\scriptsize ± 0.05\normalsize}}    & \underline{0.2630\scriptsize ± 0.02\normalsize}          &  {74.05\scriptsize ± 0.28\normalsize}                & {0.2203\scriptsize ± 0.02\normalsize}                                  & {\underline {0.0434\scriptsize ± 0.01\normalsize}}    &  {0.0844\scriptsize ± 0.01\normalsize}    & 69.80\scriptsize ± 1.48\normalsize                                          \\
\multicolumn{1}{l|}{CPA-VAE}                   &                                                                              & \underline{0.5391\scriptsize ± 0.08\normalsize}                    & 0.2085\scriptsize ± 0.11\normalsize          & 0.2408\scriptsize ± 0.03\normalsize          & 72.53\scriptsize ± 0.74\normalsize                      & \underline{0.3718\scriptsize ± 0.05\normalsize}                                        & -0.0250\scriptsize ± 0.07\normalsize         & \underline{0.1373\scriptsize ± 0.01\normalsize}          & \underline{73.00\scriptsize ± 0.44\normalsize}                                          \\
\multicolumn{1}{l|}{sVAE+}                     & \multicolumn{1}{c|}{$3\sigma$}                                               & 0.0249\scriptsize ± 0.02\normalsize                    & -0.0189\scriptsize ± 0.01\normalsize         & 0.0232\scriptsize ± 0.00\normalsize          & \underline{75.34\scriptsize ± 0.83\normalsize}                      & 0.0259\scriptsize ± 0.03\normalsize                                        & -0.0319\scriptsize ± 0.01\normalsize         & 0.0310\scriptsize ± 0.01\normalsize          & 70.77\scriptsize ± 0.74\normalsize                                          \\
\multicolumn{1}{l|}{SAMS-VAE}                  &                                                                              & 0.4594\scriptsize ± 0.03\normalsize                    & 0.2098\scriptsize ± 0.03\normalsize          & 0.2362\scriptsize ± 0.02\normalsize          & 75.18\scriptsize ± 0.61\normalsize                      & 0.0767\scriptsize ± 0.06\normalsize                                        & -0.0213\scriptsize ± 0.03\normalsize         & 0.0556\scriptsize ± 0.02\normalsize          &  {68.83\scriptsize ± 0.75\normalsize}                                    \\
\multicolumn{1}{l|}{\textbf{$\text{\modelname}_{3\sigma}$}} &                                                                              & \textbf{0.7119\scriptsize ± 0.03\normalsize}           & \textbf{0.5040\scriptsize ± 0.04\normalsize} & \textbf{0.3337\scriptsize ± 0.02\normalsize} & \textbf{93.53\scriptsize ± 0.64\normalsize}             & \textbf{0.6520\scriptsize ± 0.02\normalsize}                               & \textbf{0.3764\scriptsize ± 0.03\normalsize} & \textbf{0.4324\scriptsize ± 0.04\normalsize} & \textbf{84.83\scriptsize ± 1.59\normalsize}                                 \\ \hline
\multicolumn{1}{l|}{Conditional VAE}           &                                                                              &  {0.5396\scriptsize ± 0.04\normalsize}              & {\underline {0.2855\scriptsize ± 0.04\normalsize}}    & {\underline {0.2641\scriptsize ± 0.02\normalsize}}    &  {82.06\scriptsize ± 0.36\normalsize}                &  {0.2270\scriptsize ± 0.02\normalsize}                                  & {\underline {0.0448\scriptsize ± 0.01\normalsize}}    &  {0.0856\scriptsize ± 0.01\normalsize}    & 77.65\scriptsize ± 1.31\normalsize                                          \\
\multicolumn{1}{l|}{CPA-VAE}                   &                                                                              & \underline{0.5674\scriptsize ± 0.08\normalsize}                    & 0.2851\scriptsize ± 0.12\normalsize          & 0.2442\scriptsize ± 0.03\normalsize          & 80.16\scriptsize ± 0.77\normalsize                      & \underline{0.3845\scriptsize ± 0.05\normalsize}                                        & -0.0054\scriptsize ± 0.07\normalsize         & \underline{0.1420\scriptsize ± 0.01\normalsize}          & \underline{80.10\scriptsize ± 0.43\normalsize}                                          \\
\multicolumn{1}{l|}{sVAE+}                     & \multicolumn{1}{c|}{$4\sigma$}                                               & 0.0286\scriptsize ± 0.03\normalsize                    & -0.0185\scriptsize ± 0.01\normalsize         & 0.0230\scriptsize ± 0.00\normalsize          & 82.97\scriptsize ± 0.56\normalsize                      & 0.0220\scriptsize ± 0.03\normalsize                                        & -0.0386\scriptsize ± 0.01\normalsize         & 0.0313\scriptsize ± 0.01\normalsize          & 79.26\scriptsize ± 0.29\normalsize                                          \\
\multicolumn{1}{l|}{SAMS-VAE}                  &                                                                              & 0.4633\scriptsize ± 0.03\normalsize                    & 0.2096\scriptsize ± 0.02\normalsize          & 0.2376\scriptsize ± 0.02\normalsize          & \underline{83.20\scriptsize ± 0.69\normalsize}                      & 0.0821\scriptsize ± 0.06\normalsize                                        & -0.0220\scriptsize ± 0.03\normalsize         & 0.0565\scriptsize ± 0.02\normalsize          & { 77.04\scriptsize ± 0.78\normalsize}                                    \\
\multicolumn{1}{l|}{\textbf{$\text{\modelname}_{4\sigma}$}} &                                                                              & \textbf{0.7477\scriptsize ± 0.03\normalsize}           & \textbf{0.5423\scriptsize ± 0.04\normalsize} & \textbf{0.3620\scriptsize ± 0.02\normalsize} & \textbf{95.90\scriptsize ± 0.34\normalsize}             & \textbf{0.6572\scriptsize ± 0.03\normalsize}                               & \textbf{0.3932\scriptsize ± 0.04\normalsize} & \textbf{0.4041\scriptsize ± 0.04\normalsize} & \textbf{88.18\scriptsize ± 0.76\normalsize}                                 \\ \hline
\multicolumn{1}{l|}{Conditional VAE}           &                                                                              &  {0.5525\scriptsize ± 0.03\normalsize}              & { {0.2990\scriptsize ± 0.04\normalsize}}    & {\underline {0.2748\scriptsize ± 0.02\normalsize}}    & {{86.22\scriptsize ± 0.40\normalsize}}                & {{0.2287\scriptsize ± 0.02\normalsize}}                                  & {\underline {0.0459\scriptsize ± 0.01\normalsize}}    & {{0.0866\scriptsize ± 0.01\normalsize}}    & 81.84\scriptsize ± 1.18\normalsize                                          \\
\multicolumn{1}{l|}{CPA-VAE}                   &                                                                              & \underline{0.5814\scriptsize ± 0.08\normalsize}                    & \underline{0.3077\scriptsize ± 0.11\normalsize}          & 0.2543\scriptsize ± 0.03\normalsize          & 84.30\scriptsize ± 0.68\normalsize                      & \underline{0.3990\scriptsize ± 0.04\normalsize}                                        & 0.0274\scriptsize ± 0.06\normalsize          & \underline{0.1461\scriptsize ± 0.01\normalsize}          & \underline{83.36\scriptsize ± 0.50\normalsize}                                          \\
\multicolumn{1}{l|}{sVAE+}                     & \multicolumn{1}{c|}{$5\sigma$}                                               & 0.0298\scriptsize ± 0.03\normalsize                    & -0.0187\scriptsize ± 0.01\normalsize         & 0.0242\scriptsize ± 0.01\normalsize          & 86.88\scriptsize ± 0.49\normalsize                      & 0.0225\scriptsize ± 0.03\normalsize                                        & -0.0379\scriptsize ± 0.01\normalsize         & 0.0314\scriptsize ± 0.01\normalsize          & 83.07\scriptsize ± 0.30\normalsize                                          \\
\multicolumn{1}{l|}{SAMS-VAE}                  &                                                                              & 0.4732\scriptsize ± 0.03\normalsize                    & 0.2173\scriptsize ± 0.02\normalsize          & 0.2462\scriptsize ± 0.02\normalsize          & \underline{87.19\scriptsize ± 0.71\normalsize}                      & 0.0885\scriptsize ± 0.07\normalsize                                        & -0.0181\scriptsize ± 0.03\normalsize         & 0.0566\scriptsize ± 0.02\normalsize          & {{81.27\scriptsize ± 0.72\normalsize}}                                    \\
\multicolumn{1}{l|}{\textbf{$\text{\modelname}_{5\sigma}$}} &                                                                              & \textbf{0.7518\scriptsize ± 0.03\normalsize}           & \textbf{0.5482\scriptsize ± 0.04\normalsize} & \textbf{0.3671\scriptsize ± 0.02\normalsize} & \textbf{96.62\scriptsize ± 0.38\normalsize}             & \textbf{0.6258\scriptsize ± 0.06\normalsize}                               & \textbf{0.3239\scriptsize ± 0.05\normalsize} & \textbf{0.3493\scriptsize ± 0.07\normalsize} & \textbf{91.40\scriptsize ± 1.80\normalsize}                                 \\ \hline


\multicolumn{2}{c|}{\textbf{Dataset}}                                                                                         & \multicolumn{4}{c|}{\textbf{Replogle}}                                                                                                                                                                         & \multicolumn{4}{c}{\textbf{Adamson}}                                                                                                                                                                                                                   \\ \hline
\multicolumn{1}{c|}{Model}                     & \multicolumn{1}{c|}{\begin{tabular}[c]{@{}c@{}}QC threshold\end{tabular}} & \begin{tabular}[c]{@{}c@{}}ATE-$\rho$\end{tabular} & ATE-$R^2$                                       & Jaccard                                      & \begin{tabular}[c]{@{}c@{}}QCPR (\%)\end{tabular} & \multicolumn{1}{c}{\begin{tabular}[c]{@{}c@{}}ATE-$\rho$\end{tabular}} & \multicolumn{1}{c}{ATE-$R^2$}                   & \multicolumn{1}{c}{Jaccard}                  & \multicolumn{1}{c}{\begin{tabular}[c]{@{}c@{}}QCPR (\%)\end{tabular}} \\ \hline
\multicolumn{1}{l|}{Conditional VAE}           &                                                                              & {\underline {0.7022\scriptsize ± 0.00\normalsize}}              & {\underline {0.4883\scriptsize ± 0.01\normalsize}}    & \textbf{0.2688\scriptsize ± 0.00\normalsize} & {\underline {76.56\scriptsize ± 0.26\normalsize}}                & {\underline {0.6335\scriptsize ± 0.01\normalsize}}                                  & {\underline {0.3954\scriptsize ± 0.01\normalsize}}    & {\underline {0.3110\scriptsize ± 0.01\normalsize}}    & 77.23\scriptsize ± 0.44\normalsize                                          \\
\multicolumn{1}{l|}{CPA-VAE}                   &                                                                              & 0.5171\scriptsize ± 0.01\normalsize                    & 0.1241\scriptsize ± 0.02\normalsize          & 0.1438\scriptsize ± 0.00\normalsize          & 74.83\scriptsize ± 0.50\normalsize                      & 0.5571\scriptsize ± 0.02\normalsize                                        & 0.2637\scriptsize ± 0.03\normalsize          & 0.2123\scriptsize ± 0.01\normalsize          & 76.64\scriptsize ± 0.90\normalsize                                          \\
\multicolumn{1}{l|}{sVAE+}                     & \multicolumn{1}{c|}{$3\sigma$}                                               & 0.5780\scriptsize ± 0.01\normalsize                    & 0.3222\scriptsize ± 0.01\normalsize          & 0.1565\scriptsize ± 0.00\normalsize          & 73.89\scriptsize ± 0.53\normalsize                      & 0.5298\scriptsize ± 0.02\normalsize                                        & 0.2580\scriptsize ± 0.03\normalsize          & 0.1778\scriptsize ± 0.01\normalsize          & 76.27\scriptsize ± 0.98\normalsize                                          \\
\multicolumn{1}{l|}{SAMS-VAE}                  &                                                                              & 0.6798\scriptsize ± 0.03\normalsize                    & 0.4584\scriptsize ± 0.04\normalsize          & 0.2404\scriptsize ± 0.02\normalsize          & 74.96\scriptsize ± 0.69\normalsize                      & 0.3901\scriptsize ± 0.01\normalsize                                        & 0.1432\scriptsize ± 0.01\normalsize          & 0.1846\scriptsize ± 0.01\normalsize          & {\underline {77.34\scriptsize ± 0.78\normalsize}}                                    \\
\multicolumn{1}{l|}{\textbf{$\text{\modelname}_{3\sigma}$}} &                                                                              & \textbf{0.7192\scriptsize ± 0.01\normalsize}           & \textbf{0.5155\scriptsize ± 0.01\normalsize} & {\underline {0.2667\scriptsize ± 0.01\normalsize}}    & \textbf{97.33\scriptsize ± 0.04\normalsize}             & \textbf{0.7529\scriptsize ± 0.01\normalsize}                               & \textbf{0.5611\scriptsize ± 0.02\normalsize} & \textbf{0.3471\scriptsize ± 0.01\normalsize} & \textbf{89.92\scriptsize ± 0.47\normalsize}                                 \\ \hline
\multicolumn{1}{l|}{Conditional VAE}           &                                                                              & {\underline {0.7255\scriptsize ± 0.01\normalsize}}              & {\underline {0.5233\scriptsize ± 0.01\normalsize}}    & {\underline {0.2776\scriptsize ± 0.00\normalsize}}    & {\underline {84.36\scriptsize ± 0.24\normalsize}}                & {\underline {0.6435\scriptsize ± 0.01\normalsize}}                                  & {\underline {0.4059\scriptsize ± 0.02\normalsize}}    & {\underline {0.3109\scriptsize ± 0.01\normalsize}}    & 85.06\scriptsize ± 0.52\normalsize                                          \\
\multicolumn{1}{l|}{CPA-VAE}                   &                                                                              & 0.5352\scriptsize ± 0.01\normalsize                    & 0.1765\scriptsize ± 0.03\normalsize          & 0.1494\scriptsize ± 0.01\normalsize          & 82.92\scriptsize ± 0.38\normalsize                      & 0.5715\scriptsize ± 0.02\normalsize                                        & 0.2863\scriptsize ± 0.03\normalsize          & 0.2103\scriptsize ± 0.01\normalsize          & 84.80\scriptsize ± 0.67\normalsize                                          \\
\multicolumn{1}{l|}{sVAE+}                     & \multicolumn{1}{c|}{$4\sigma$}                                               & 0.6056\scriptsize ± 0.01\normalsize                    & 0.3612\scriptsize ± 0.01\normalsize          & 0.1661\scriptsize ± 0.00\normalsize          & 82.15\scriptsize ± 0.50\normalsize                      & 0.5437\scriptsize ± 0.02\normalsize                                        & 0.2774\scriptsize ± 0.03\normalsize          & 0.1773\scriptsize ± 0.01\normalsize          & 84.66\scriptsize ± 0.60\normalsize                                          \\
\multicolumn{1}{l|}{SAMS-VAE}                  &                                                                              & 0.7086\scriptsize ± 0.03\normalsize                    & 0.4941\scriptsize ± 0.04\normalsize          & 0.2516\scriptsize ± 0.02\normalsize          & 83.01\scriptsize ± 0.43\normalsize                      & 0.3939\scriptsize ± 0.01\normalsize                                        & 0.1442\scriptsize ± 0.01\normalsize          & 0.1808\scriptsize ± 0.01\normalsize          & {\underline {85.36\scriptsize ± 0.75\normalsize}}                                    \\
\multicolumn{1}{l|}{\textbf{$\text{\modelname}_{4\sigma}$}} &                                                                              & \textbf{0.7565\scriptsize ± 0.01\normalsize}           & \textbf{0.5595\scriptsize ± 0.01\normalsize} & \textbf{0.2869\scriptsize ± 0.01\normalsize} & \textbf{98.10\scriptsize ± 0.18\normalsize}             & \textbf{0.7636\scriptsize ± 0.01\normalsize}                               & \textbf{0.5770\scriptsize ± 0.01\normalsize} & \textbf{0.3367\scriptsize ± 0.01\normalsize} & \textbf{93.66\scriptsize ± 0.56\normalsize}                                 \\ \hline
\multicolumn{1}{l|}{Conditional VAE}           &                                                                              & {\underline {0.7296\scriptsize ± 0.01\normalsize}}              & {\underline {0.5282\scriptsize ± 0.01\normalsize}}    & {\underline {0.2793\scriptsize ± 0.00\normalsize}}    & {\underline {88.45\scriptsize ± 0.27\normalsize}}                & {\underline {0.6484\scriptsize ± 0.01\normalsize}}                                  & {\underline {0.4110\scriptsize ± 0.02\normalsize}}    & {\underline {0.3110\scriptsize ± 0.01\normalsize}}    & 88.75\scriptsize ± 0.45\normalsize                                          \\
\multicolumn{1}{l|}{CPA-VAE}                   &                                                                              & 0.5380\scriptsize ± 0.02\normalsize                    & 0.1999\scriptsize ± 0.03\normalsize          & 0.1501\scriptsize ± 0.01\normalsize          & 87.20\scriptsize ± 0.38\normalsize                      & 0.5758\scriptsize ± 0.02\normalsize                                        & 0.2928\scriptsize ± 0.04\normalsize          & 0.2102\scriptsize ± 0.01\normalsize          & 88.34\scriptsize ± 0.57\normalsize                                          \\
\multicolumn{1}{l|}{sVAE+}                     & \multicolumn{1}{c|}{$5\sigma$}                                               & 0.6137\scriptsize ± 0.01\normalsize                    & 0.3736\scriptsize ± 0.01\normalsize          & 0.1694\scriptsize ± 0.00\normalsize          & 86.60\scriptsize ± 0.41\normalsize                      & 0.5488\scriptsize ± 0.02\normalsize                                        & 0.2843\scriptsize ± 0.03\normalsize          & 0.1776\scriptsize ± 0.01\normalsize          & 88.65\scriptsize ± 0.55\normalsize                                          \\
\multicolumn{1}{l|}{SAMS-VAE}                  &                                                                              & 0.7167\scriptsize ± 0.03\normalsize                    & 0.4998\scriptsize ± 0.03\normalsize          & 0.2558\scriptsize ± 0.02\normalsize          & 87.26\scriptsize ± 0.33\normalsize                      & 0.3952\scriptsize ± 0.01\normalsize                                        & 0.1442\scriptsize ± 0.01\normalsize          & 0.1792\scriptsize ± 0.01\normalsize          & {\underline {89.05\scriptsize ± 0.49\normalsize}}                                    \\
\multicolumn{1}{l|}{\textbf{$\text{\modelname}_{5\sigma}$}} &                                                                              & \textbf{0.7638\scriptsize ± 0.01\normalsize}           & \textbf{0.5719\scriptsize ± 0.01\normalsize} & \textbf{0.2931\scriptsize ± 0.01\normalsize} & \textbf{98.41\scriptsize ± 0.13\normalsize}             & \textbf{0.7609\scriptsize ± 0.01\normalsize}                               & \textbf{0.5723\scriptsize ± 0.01\normalsize} & \textbf{0.3153\scriptsize ± 0.00\normalsize} & \textbf{94.34\scriptsize ± 0.42\normalsize}                                 \\ \hline

\end{tabular}
}
}
\captionsetup{justification=raggedright, singlelinecheck=false}
\caption{Quantitative evaluation on Norman dataset, Dixit dataset, Replogle dataset and Adamson dataset across ${3\sigma, 4\sigma, 5\sigma}$ quality control (QC) thresholds. Note that the QC threshold column refers to the cut-off point -- defined as delta-MAD threshold -- of the generated data to be included in the evaluation phase. Best results are in bold-faced while second-best ones are underlined.}
\label{tab:main_results}

\end{table*}

\begin{table}[hbt!]
{\Huge
\resizebox*{0.6\columnwidth}{!}{%
\renewcommand{\arraystretch}{1.3}
\begin{tabular}{l|l|cccc}
\hline
\textbf{Model}                  & \multicolumn{1}{c|}{\begin{tabular}[c]{@{}c@{}}QC thr.\end{tabular}} & \begin{tabular}[c]{@{}c@{}}ATE-$\rho$\end{tabular} & ATE-$R^2$                                       & Jaccard                                      & \begin{tabular}[c]{@{}c@{}}QCPR (\%)\end{tabular} \\ \hline
$\text{\modelname}_{3\sigma}$             &                                                                              & \textbf{0.7119\large ± 0.03\large}           & \textbf{0.5040\large ± 0.04\large} & \textbf{0.3337\large ± 0.02\large} & \textbf{93.53\large ± 0.64\large}             \\
$\text{\modelname}_{3\sigma}$ ~w/o CF     & \multicolumn{1}{c|}{$3\sigma$}                                               & 0.6505\large ± 0.02\large                    & 0.4210\large ± 0.02\large          & {\underline {0.3046\large ± 0.01\large}}    & 91.46\large ± 0.73\large                      \\
$\text{\modelname}_{3\sigma}$ ~w/o Causal &                                                                              & {\underline {0.7018\large ± 0.02\large}}              & {\underline {0.4844\large ± 0.02\large}}    & 0.2938\large ± 0.01\large          & {\underline {92.63\large ± 0.71\large}}                \\ \hline
$\text{\modelname}_{4\sigma}$             &                                                                              & \textbf{0.7477\large ± 0.03\large}           & \textbf{0.5423\large ± 0.04\large} & \textbf{0.3620\large ± 0.02\large} & \textbf{95.90\large ± 0.34\large}             \\
$\text{\modelname}_{4\sigma}$ ~w/o CF     & \multicolumn{1}{c|}{$4\sigma$}                                               & {\underline {0.7111\large ± 0.03\large}}              & {\underline {0.4927\large ± 0.04\large}}    & {\underline {0.3240\large ± 0.01\large}}    & {\underline {94.24\large ± 0.43\large}}                \\
$\text{\modelname}_{4\sigma}$ ~w/o Causal &                                                                              & 0.7058\large ± 0.03\large                    & 0.4790\large ± 0.05\large          & 0.2946\large ± 0.01\large          & 87.90\large ± 5.34\large                      \\ \hline
$\text{\modelname}_{5\sigma}$             &                                                                              & \textbf{0.7518\large ± 0.03\large}           & \textbf{0.5482\large ± 0.04\large} & \textbf{0.3671\large ± 0.02\large} & \textbf{96.62\large ± 0.38\large}             \\
$\text{\modelname}_{5\sigma}$ ~w/o CF     & \multicolumn{1}{c|}{$5\sigma$}                                               & {\underline {0.7395\large ± 0.02\large}}              & {\underline {0.5315\large ± 0.03\large}}    & {\underline {0.3540\large ± 0.01\large}}    & {\underline {95.71\large ± 0.49\large}}                \\
$\text{\modelname}_{5\sigma}$ ~w/o Causal &                                                                              & 0.6875\large ± 0.03\large                    & 0.4402\large ± 0.05\large          & 0.3008\large ± 0.03\large          & 92.85\large ± 4.06\large                      \\ \hline
\end{tabular}
\vspace{1em}
}
}
\vspace{1em}
\captionsetup{justification=raggedright, singlelinecheck=false}
\caption{Experimental results on ablated versions of $\text{\modelname}_{\sigma}$. Best results are in bold-faced while second-best ones are underlined.}
\label{tab:ablation_results}

\end{table}

\subsection{Experimental Results}
Table~\ref{tab:main_results} shows the quantitative results on the four Perturb-seq datasets. According to the results, \modelname~overall surpassed all of its baselines in the three evaluation metrics that measure the model's ability to accurately predict cellular responses. Moreover, \modelname~achieved the highest QC Pass Rate across all datasets and QC threshold settings, demonstrating its ability to capture the true data distribution of QC passed gene expression profiles due to additional disentanglement of latent artifacts during its training phase. Notably, despite multi-gene perturbation cellular response prediction being more challenging than that of single-gene perturbation, \modelname~significantly outperforms the second-best model with a large margin, particularly in the Norman and Dixit datasets, both of which contain multi-gene perturbation scRNA-seq data. This highlights \modelname's strong generalizability in out-of-distribution (OOD) gene perturbation treatment scenarios. 

\subsection{Ablation Study}
To investigate the effects of utilizing causal distribution of artifact disentanglement and our proposed auxiliary loss objective utilizing counterfactual reasoning related to presence of technical artifacts, we conducted experiments on the ablated versions of \modelname~which are denoted as \modelname~w/o Causal and \modelname~w/o CF respectively. The former models the technical artifact as fixed learnable embedding instead of parameterized prior distribution (Algorithm~\ref{alg:encoding}.5). The latter removes the KL divergence-based auxiliary loss objective, eliminating the counterfactual reasoning-based approach in aligning the latent basal state embeddings ($\mathcal{J}_2$).

As shown in Table~\ref{tab:ablation_results}, the ablated versions of \modelname~exhibited performance decline, implying the benefits of employing counterfactual reasoning and causal learning. Particularly, we find that modeling the technical artifact as a learnable embedding (\modelname~w/o Causal) results in a sharper decline, especially at the 5$\sigma$ QC threshold. While setting a higher QC threshold leads to imbalance between the number of QC passed and failed samples, we speculate that distribution-based artifact modeling is more resilient to such issues compared to its embedding-based version. The effect of removing the counterfactual reasoning (\modelname~w/o CF) is more profound at the 3$\sigma$ threshold. This outcome aligns with our assumption that the KL loss objective between the counterfactual latent basal state embeddings aids in the learning of artifact features, particularly when generalization is well-established due to the balanced data instances.

\begin{figure*}[!t]
\centering
    \includegraphics[width=\textwidth]{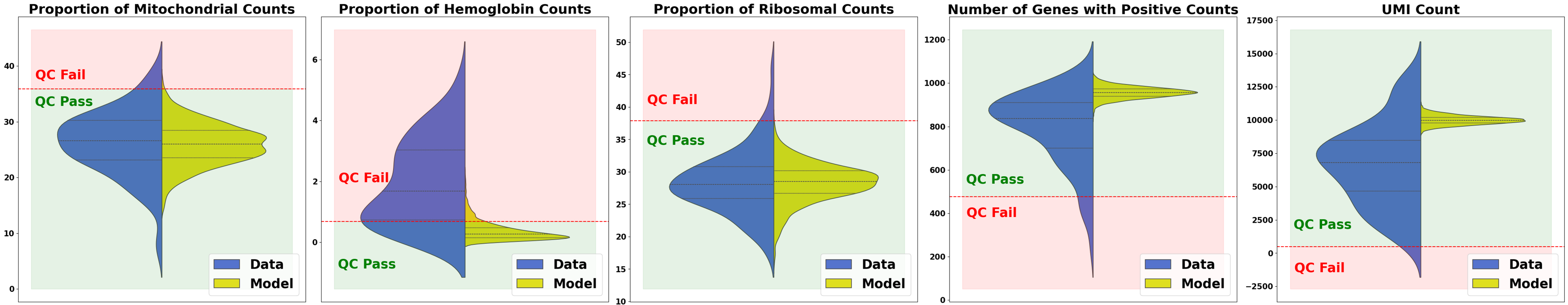}
\caption{Violin plots showing the data(blue) and model-generated(green) distribution of POLD3-perturbed cellular response for each QC sub-criteria. The red dotted line refers to the predefined QC threshold, with the green-colored region representing QC passed values and the red-colored region representing QC failed values.}
\label{fig:analysis_violinplot}
\end{figure*}

\begin{figure*}[!t]
\centering
    \includegraphics[width=1\textwidth]{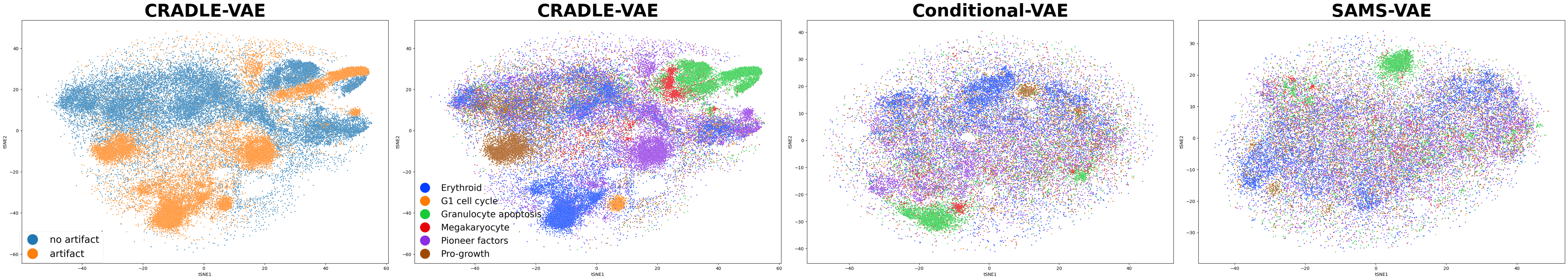}
\caption{t-SNE plots labelled by the presence of artifacts (left 1) and  by perturbation types (right 3) for \modelname, conditional-VAE, and SAMS-VAE, respectively.}
\label{fig:analysis_tsne}
\end{figure*}

\subsection{Distributional Generative Quality Analysis}
To further analyze \modelname's generative quality, we visualized the distributions of actual (Replogle) and model-generated gene counts related to the QC criteria, that results from a specific treatment perturbing the POLD3 gene~\citep{replogle2022mapping}. The rationale behind selecting this particular perturbation is as follows: 1) the number of gene expression profiles treated by this perturbation in the dataset is relatively low (85 compared to the average of 164), 2) only 13\% of them passes the QC criteria. This may pose challenges in learning the causal distributions during the training process, especially if the latent effects of technical artifacts are not properly addressed. We expect these challenges to be dealt with the employment of counterfactual reasoning-based artifact disentanglement. Figure~\ref{fig:analysis_violinplot} shows that \modelname~exhibits its consistency in robustly generating read counts that satisfy all QC sub-criteria.

We move our focus to a critical sub-criterion responsible for a significant decline in data quality. The distribution of hemoglobin counts in the Replogle dataset predominantly exceed the QC threshold, leading to a high QC failure rate. On the contrary, the distribution generated by \modelname~is shifted below the threshold, implying a marked enhancement in generative data quality. For both the number of genes with positive counts and UMI count, the violin plots in Figure~\ref{fig:analysis_violinplot} display a skewed distribution compared to the original data, indicating that \modelname's generated gene expression profiles yield consistent and higher quality outcomes.

\subsection{Disentanglement Effect Analysis}
We investigated the effects of \modelname's disentanglement of two important variables which are perturbation and artifact effects. We utilized t-SNE in visualizing the high-dimensional gene expression profiles generated by \modelname, and colored them based on which pathway clusters are relevant to each of their gene perturbations. This aligns with a domain-specific assertion stating that perturbation of genes with similar biological roles are expected to show similar expression patterns. Following the method in~\citep{norman2019exploring}, we grouped them into six pathways for this visualization. As illustrated in Figure~\ref{fig:analysis_tsne}, \modelname~appears to form clearer clusters within the same pathway compared to other models, particularly for those related to the pro-growth and megakaryocyte pathways.

Additionally, we examined the disentanglement of artifacts by comparing the same generated data with and without artifacts. In Figure~\ref{fig:analysis_tsne}, the t-SNE visualization within pathways shows distinct clustering based on the presence or absence of artifacts, suggesting that our model successfully disentangles artifact effects. Overall, these findings suggest that our model can meaningfully separate both latent perturbation and artifact variables, as reflected by the well-defined clusters in the visualizations.

\section{Conclusion}
Quality issues in scRNA-seq datasets have been overlooked despite the improvements in predicting cellular responses achieved by previous works. We propose a causal inference-based VAE model \modelname~which has several advantages. During the training process, \modelname~disentangles not only latent perturbation effects but also the artifacts that inherently degrade data quality. Additionally, the disentanglement of these artifacts is further enhanced by our novel counterfactual reasoning-based approach which employs an auxiliary loss objective used for aligning the counterfactual basal states. As demonstrated in our experiments and analysis, \modelname~is capable of accurately predicting cellular responses with improved generative quality. We expect that \modelname~addresses the quality issues of both experimentally measured and model-generated single-cell response data upon gene perturbation, eliminating the need of arbitrary quality control standards for scRNA-seq data analysis.

\section{Acknowledgement}
This research was supported by the National Research Foundation of Korea [NRF2023R1A2C3004176, RS-2023-00262002], the Ministry of Health \& Welfare, Republic of Korea [HR20C0021(3)], ICT Creative Consilience Program through the Institute of Information \& Communications Technology Planning \& Evaluation (IITP) grant funded by the Korea government (MSIT) [IITP-2024- 20200-01819].

This work was supported by Hankuk University of Foreign Studies Research Fund (of 2024).

Figure~\ref{figure:cradlevae} was created with BioRender.com.

\newpage
\bibliographystyle{hunsrtnat}

\maketitle
\def\modelname{\textsc{Cradle-VAE}}

\clearpage
\onecolumn

\appendix
\renewcommand{\thesection}{\Alph{section}.\arabic{section}}
\setcounter{section}{0}

\begin{appendices}
 
\section{List of Notations}
\begin{tabbing}
    \hspace{2cm} \= \kill 
    \( N \) \> number of data samples\\
    \( X \) \> gene expression matrix ($\in \mathbb R^{N\times D_x}$)\\
    \( x_i \) \>  gene expression vector of sample $i$ ($\in \mathbb R^{D_x}$)\\
    \( P \) \> gene perturbation matrix ($\in \left\{ 0,1 \right\}^{N\times T}$)\\
    \( p_i \) \> gene perturbation vector of sample $i$ ($\in \left\{ 0,1 \right\}^{T}$), 1 if gene $t$ is perturbed\\
    \( A \) \> artifact presence labels ($\in \left\{ 0,1 \right\}^N$)\\
    \( a_i \) \> artifact presence label of sample $i$ ($\in \left\{ 0,1 \right\}$), 1 if artifact is present\\
    \( D_x \) \> total number of genes\\
    \( T \) \> number of perturbation types\\
    \( \mathbf{z}^b_i \) \> latent basal state embedding of sample $i$ ($\in \mathbb R^{D_z}$)\\
    \( \mathbf{z}^p_i \) \> latent perturbation effect embedding of sample $i$ ($\in \mathbb R^{D_z}$)\\
    \( \mathbf{z}^a_i \) \> latent artifact embedding of sample $i$ ($\in \mathbb R^{D_z}$)\\
    \( Z^b \) \> latent basal state embeddings ($\in \mathbb R^{N\times D_z}$)\\
    \( D_z \) \> dimension size of latent subspaces\\
    \( E \) \> global latent perturbation embeddings ($\in \mathbb R^{T\times D_z}$)\\
    \( e_t \) \> global gene-wise perturbation effects ($\in \mathbb R^{D_z}$)\\
    \( M \) \> global latent perturbation masks ($\in \left\{0,1\right\}^{T\times D_z}$)\\
    \( m_t \) \> global sparse latent offsets ($\in \left\{0,1\right\}^{D_z}$)\\
    \( \hat{\omega}_t \) \> learnable parameter\\
    \( U \) \> global latent artifact embeddings ($\in \mathbb R^{1\times D_z}$)\\
    \( \mathbf{u} \) \> global latent artifact embedding ($\in \mathbb R^{D_z}$)\\ 
    \( \hat{\mu},\hat{\sigma}\) \> learnable parameters\\
    \( 1_t \) \> one-hot encoding of the $t$th gene perturbation treatment\\
    \( \hat{f}_{emb} \) \> trainable neural network of perturbation\\
    \( \hat{f}_{enc} \) \> trainable neural network of basal state\\
    \( \tilde{x} \) \> generated gene expression profile\\
    \( \hat{f}_{dec} \) \> learnable neural network with softmax output\\
    \( \hat{f}_{dec}(z_i) \) \> frequency of each transcript in sample $i$ ($\in [0,1]^{D_x}$)\\
    \( l_i \) \> total number of read counts in sample $i$\\
    \( \theta_d \) \> learnable inverse dispersion used universally across all cells\\
    \( \oplus \) \> vector concatenation\\
    \( \bullet \) \> Hadamard product operation\\
    \( \otimes \) \> matrix multiplication operation\\
    \( \mathcal{N} \) \> Normal Distribution\\
    \( \text{Bernoulli} \) \> Bernoulli Distribution\\
    \( \Gamma-\text{Poisson} \) \> Gamma-Poisson Distribution\\
    \( \mathbf{z}_{i,c}^a \) \> counterfactual latent artifact embedding of sample $i$\\
    \( \mathbf{z}_{i,c}^b \) \> counterfactual latent basal state embedding of sample $i$\\
    \( \bar{x}_{i,c} \) \> median of sampled counterfactuals of sample $i$\\
    \( \bar{\mathbf{z}}_{i,c}^b \) \> reference counterfactual latent basal state embedding of sample $i$\\
    \( \phi \) \> learnable parameters of encoder\\
    \( \theta \) \> learnable parameters of decoder\\
    \( q \) \> encoder\\
    \( p \) \> decoder\\

\end{tabbing}

\newpage
\section{Baselines}
\subsection{SAMS-VAE}
SAMS-VAE is a fully defined VAE-based generative model designed modeling perturbation effects in cells\citep{bereket2024modelling}. Similar to \modelname, SAMS-VAE specifies prior probability distributions for both the latent perturbation effects and latent basal state. While it also incorporates sparsity in the latent perturbation effects, it does not explicitly address or model the technical artifacts present in the data, thus requiring preprocessing of the scRNA-seq training data.

\subsection{SVAE+}
SVAE+ also explicitly addresses sparsity in the data using a mask and embedding mechanism~\citep{lopez2023learning}. However, unlike \modelname, SVAE+ lacks a mechanism for composing multiple interventions. Instead, it operates by sampling a cell's full latent embedding from a learned prior, which is conditioned on the treatment the cell receives. The differences in how SVAE+ handles the cell's latent state and the variational inference methods it employs set it apart from our model.

\subsection{CPA-VAE}
CPA-VAE is the ablated model of SAMS-VAE defined by \citeauthor{bereket2024modelling} that is identical to SAMS-VAE with all mask components fixed to 1. In another words, it does not incorporate sparsity to the latent perturbation effects. However, it inherits the benefits of the inference improvements to the correlated variational families.

\subsection{Conditional VAE}
Conditional VAE ia a deep conditional generative model initially proposed for structured output predictions .~\citep{sohn2015learning}. It adopted stochastic neural networks for the task based on the generative model with Gaussian latent variables. We chose it as baselines because it shares the same characteristics like VAE backbone and the incorporation of input omission noise in the reconstruction process to regularize the deep neural networks during training.

\section{Concept Description}
\subsection{Perturb-seq}
Perturb-seq is a technique that combines CRISPR-based gene perturbation with single-cell RNA sequencing (scRNA-seq). Perturb-seq combines the flexibility of CRISPR/Cas9 for targeting one or multiple genes with the large-scale capabilities of scRNA-seq to generate comprehensive genomic data. This technique has been applied in both post-mitotic immune cells and proliferating cell lines, allowing researchers to examine how genetic perturbations influence gene expression and cell states at a single-cell level.

\subsection{Causal Inference}
In the context of machine learning, causal inference is a method used to understand and model the cause-and-effect relationships between variables rather than just their correlations~\citep{spirtes2010introduction}. Traditional machine learning models focus on finding patterns in data, but these correlations may be influenced by other variables and do not always represent true causal links. Causal inference addresses this limitation by using methods like causal discovery to learn causal graphs and causal effect estimation to quantify the impact of interventions. 
Causal modeling is divided into three stages: 1) associational causality, which predicts in the i.i.d. setting, 2) interventional causality, which predicts under distribution shifts, and 3) counterfactual causality, which answers counterfactual questions and serves as the main concept we apply in \modelname's methodology.

\subsection{Counterfactual Reasoning}
Counterfactual reasoning attempts to answer the question of what the model would predict if the action had been different. In machine learning, it involves estimating the probable outcomes that could have occurred if treatment B were taken instead of treatment A. The concept of counterfactual reasoning is particularly relevant in understanding causal relationships. In this paper, we use counterfactual reasoning to ask the counterfactual question: \textit{What would the outcome have been if the outcome had not contained technical artifacts, given a treatment (perturbation)?}

\subsection{Quality Control Criteria}
The six quality control sub-criteria mentioned in the main text are based on the analysis guides provided by 10X Genomics~\citep{analysisguide2022tenex}. The detailed descriptions of each are as follows:
\begin{itemize}
    \item \textbf{UMI counts} refer to the number of Unique Molecular Identifiers (UMIs) detected for each cell in single-cell RNA sequencing (scRNA-seq) experiments. UMIs are short, unique sequences added to each RNA molecule during the library preparation process. Filtering cell barcodes with too few UMIs can reduce noise and improve the accuracy of the data.
    \item \textbf{Number of features} refers to the number of distinct genes or transcripts detected in a single cell. Excluding barcodes with unusually high or low numbers of features helps remove potential multiplets or droplets with ambient RNAs. Like UMI counts, thresholds can be set arbitrarily or based on statistical measures. A high number of features may indicate that a cell is expressing a wide range of genes, which might be expected in healthy, viable cells. Cells with a low number of features might not represent viable cells and are therefore conventionally excluded in the filtering process.
    \item \textbf{Percent of mitochondrial (mt)} reads refers to the RNA transcripts originating from mitochondrial DNA that are captured and sequenced during the experiment. Cells with high mitochondrial RNA levels may be unhealthy or damaged. 
    \item \textbf{Percent of hemoglobin (hb) reads} refers to RNA transcripts associated with hemoglobin genes, which are involved in oxygen transport in red blood cells. In non-hematopoietic tissues or experiments where red blood cells are not the focus, a high proportion of hemoglobin reads can be a sign of contamination or an issue with sample preparation. 
    \item \textbf{Percent of ribosomal (rb) reads} refers to the proportion of sequencing reads that originate from ribosomal RNA (rRNA) in an RNA-seq dataset. A high percentage of ribosomal reads could indicate that the rRNA depletion step was ineffective. A high proportion of rRNA can dominate the sequencing data, reducing the amount of useful data for analyzing gene expression.
    \item \textbf{Doublets} in scRNA-seq refer to artifacts that occur when two or more cells are captured together in a single droplet or well during the sequencing process. Doublets need to be excluded because they can lead to misleading results, as the combined gene expression profiles from multiple cells can mimic the expression patterns of a single cell type or create hybrid profiles that do not represent any real biological cell state.
\end{itemize}

\section{Dataset}
\subsection{Norman dataset}
The Norman dataset includes gene expression profiles from the K562 leukemia cell line subjected to CRISPR activation (CRISPRa). The original dataset from ~\citet{norman2019exploring} is publicly available from GEO (GSE133344). For our experiment, we downloaded the processed data provided by ~\citet{ji2021machine}, and followed their preprocessing step. The preprocessed data included 111,255 cells and 19,018 genes, encompassing 131 multi-gene perturbations and 105 single-gene perturbations, with each perturbation containing approximately 300–700 samples.
 
\subsection{Dixit dataset}
The Dixit dataset contains gene expression profiles from the K562 leukemia cell line perturbed by CRISPR-Cas9 KO. The original dataset from ~\citet{dixit2016perturb} is publicly available from GEO (GSE90063). For our experiment, we used the processed data from ~\citet{ji2021machine}, and followed their preprocessing step. The preprocessed data included 103,420 cells and 18,531 genes, with 45 multi-gene perturbations and 10 single-gene perturbations, where number of samples for single-gene perturbations ranged from 4000 to 27000 and multi-gene perturbation samples contained about 60-400 samples.

\subsection{Replogle dataset}
The Replogle dataset contains genome-wide perturbations of the K562 leukemia cell line with CRISPR interference (CRISPRi). The original dataset from ~\citet{replogle2022mapping} is publicly available from the original paper. From the raw data containing 1,989,578 cells with 9,867 perturbations, we preprocessed the data following  ~\citet{lopez2023learning}, which resulted 118,641 cells and 1,187 genes, with 722 single-gene perturbations. Each perturbation contained 20-2000 samples, with mean 144 and median 164.  

\subsection{Adamson dataset}
The Adamson dataset includes gene expression data from the K562 leukemia cell line with CRISPR interference (CRISPRi). The original dataset from ~\citet{adamson2016multiplexed} is publicly available from GEO (GSE90546). We downloaded the processed data from ~\citet{peidli2024scperturb}, and followed the preprocessing step from ~\citet{ji2021machine}, which resulted 62,623 cells and 17,115 genes, with 87 unique single-gene perturbations, each replicated in approximately 100 cells. 

\section{Proof of Theorem}

\subsection{Derivation of ELBO}
The Evidence Lower Bound (ELBO) is derived from the marginal likelihood \( p(X \mid P, A) \). 
First, recall that the log marginal likelihood can be expressed as:

\[
\log p(X \mid P, A) = \log \int_{Z^b, M, E, U} p(X, Z^b, M, E, U \mid P, A) \, dZ^b \, dM \, dE \, dU.
\]

To simplify this, we introduce a variational distribution \( q(Z^b, M, E, U \mid X, P, A; \phi) \) and apply Jensen's inequality:

\[
\log p(X \mid P, A) \geq \mathbb{E}_{q(Z^b, M, E, U \mid X, P, A; \phi)} \left[ \log \frac{p(X, Z^b, M, E, U \mid P, A)}{q(Z^b, M, E, U \mid X, P, A; \phi)} \right].
\]

Here, the ELBO \(\mathcal{J}_1(\theta, \phi)\) is defined as:

\[
\mathcal{J}_1(\theta, \phi) = \mathbb{E}_{q(Z^b, M, E, U \mid X, P, A; \phi)} \left[ \log \frac{p(X, Z^b, M, E, U \mid P, A; \theta)}{q(Z^b, M, E, U \mid X, P, A; \phi)} \right].
\]

Expanding the expectation:

\begin{align}
\mathcal{J}_1(\theta, \phi) &= \mathbb{E}_{q(Z^b, M, E, U \mid X, P, A; \phi)} \left[ \log p(X, Z^b, M, E, U \mid P, A; \theta) - \log q(Z^b, M, E, U \mid X, P, A; \phi) \right] \\
&= \mathbb{E}_{q(Z^b, M, E, U \mid X, P, A; \phi)} \left[ \log p(X, Z^b, M, E, U \mid P, A; \theta) \right] - \mathbb{E}_{q(Z^b, M, E, U \mid X, P, A; \phi)} \left[ \log q(Z^b, M, E, U \mid X, P, A; \phi) \right].
\end{align}

This ELBO provides a lower bound on the log marginal likelihood \( \log p(X \mid P, A) \), which is useful for optimizing the variational parameters \(\phi\) and model parameters \(\theta\) in variational inference.
\\

\subsection{Proof Using Variational Causal Inference}

We aim to minimize the Kullback–Leibler (KL) divergence between two variational distributions , as given by the following auxiliary loss objective:
\begin{align}
    \text{KL}\left[ q(Z^b_c | X, P, A; \phi) \| q(\bar{Z}^b_c | \bar{X}, P, A; \phi) \right] &= \mathbb{E}_{q(Z^b_c | X, P, A; \phi)} \left[ \log \frac{q(Z^b_c | X, P, A; \phi)}{q(\bar{Z}^b_c | \bar{X}, P, A; \phi)} \right] \label{eq:kl}
\end{align}
where $Z^b_c$ is the counterfactual latent basal state, $\bar{X}$ is the reference counterfactual latent basal state, $P$ is the gene perturbation, $A$ is the artifact presence, $\phi$ represents the encoder learnable parameters.

The goal is to minimize this KL divergence. Start by considering the expected log-likelihood of the latent variable $Z^b_c$ under the distribution $q(Z^b_c | X, P, A; \phi)$:
\begin{align}
    \log p(\bar{Z}^b_c | \bar{X}, P, A) &= \log \mathbb{E}_{q(Z^b_c | X, P, A; \phi)} \left[ \frac{p(\bar{Z}^b_c | Z^b_c, \bar{X}, P, A)}{p(Z^b_c | X, P, A)} \right] \\
    &\geq \mathbb{E}_{q(Z^b_c | X, P, A; \phi)} \left[ \log \frac{p(\bar{Z}^b_c | Z^b_c, \bar{X}, P, A)}{p(Z^b_c | X, P, A)} \right]    \text{(Jensen's inequality)} \\
    &\geq \mathbb{E}_{q(Z^b_c | X, P, A; \phi)} \left[ \log \frac{p(\bar{Z}^b_c | Z^b_c, \bar{X}, P, A) \cdot p(Z^b_c | X, P, A)}{p(Z^b_c | X, P, A) \cdot q(Z^b_c | X, P, A; \phi)} \right] \\
    &= \mathbb{E}_{q(Z^b_c | X, P, A; \phi)} \left[ \log p(\bar{Z}^b_c | Z^b_c, \bar{X}, P, A) \right] - \text{KL}\left[ q(Z^b_c | X, P, A; \phi) \| p(Z^b_c | X, P, A) \right] \\
    &\geq \mathbb{E}_{q(Z^b_c | X, P, A; \phi)} \left[ \log p(\bar{Z}^b_c | Z^b_c, \bar{X}, P, A) \right] - \text{KL}\left[ q(Z^b_c | X, P, A; \phi) \| q(\bar{Z}^b_c | \bar{X}, P, A; \phi) \right] 
\end{align}

Rearranging the equation, we get: 
\begin{align}
    \log p(\bar{Z}^b_c | \bar{X}, P, A) + \text{KL}\left[ q(Z^b_c | X, P, A; \phi) \| q(\bar{Z}^b_c | \bar{X}, P, A; \phi) \right] &\geq \mathbb{E}_{q(Z^b_c | X, P, A; \phi)} \left[ \log p(\bar{Z}^b_c | Z^b_c, \bar{X}, P, A) \right] \label{eq:final_form}
\end{align}

Minimizing the KL divergence term $\text{KL}\left[ q(Z^b_c | X, P, A; \phi) \| q(\bar{Z}^b_c | \bar{X}, P, A; \phi) \right]$ ensures that the latent embeddings $Z^b_c$ align with $\bar{Z}^b_c$.

\newpage
\section{Results}
Additional results that were not shown in the main paper are included in this section. Quantitative evaluation on top 20 results of ATE-$\rho$, ATE-$R^2$, and Jaccard are shown in Table 1. Also, proof of concept to check the data qualtiy-quantity tradeoff is done in Table 2.

\begin{table*}[hbt!]
{\large
\resizebox*{\textwidth}{!}{%
\renewcommand{\arraystretch}{1.15}

\begin{tabular}{ll|ccc|lll}
\hline
\multicolumn{2}{c|}{\textbf{Dataset}}                                                           & \multicolumn{3}{c|}{\textbf{Norman}}                                                                                                       & \multicolumn{3}{c}{\textbf{Dixit}}                                                                                                         \\ \hline
\multicolumn{1}{c|}{Model}                                  & \multicolumn{1}{c|}{QC threshold} & ATE-$\rho$ Top 20                            & ATE-$R^2$ Top 20                             & Jaccard Top 20                               & \multicolumn{1}{c}{ATE-$\rho$ Top 20}        & \multicolumn{1}{c}{ATE-$R^2$ Top 20}         & \multicolumn{1}{c}{Jaccard Top 20}           \\ \hline
\multicolumn{1}{l|}{Conditional VAE}                        &                                   & {\underline {0.8032\scriptsize ± 0.03\normalsize}}    & {\underline {0.6301\scriptsize ± 0.05\normalsize}}    & {\underline {0.3170\scriptsize ± 0.03\normalsize}}    & 0.7051\scriptsize ± 0.07\normalsize          & 0.1410\scriptsize ± 0.02\normalsize          & 0.0943\scriptsize ± 0.03\normalsize          \\
\multicolumn{1}{l|}{CPA-VAE}                                &                                   & 0.7662\scriptsize ± 0.06\normalsize          & 0.5365\scriptsize ± 0.11\normalsize          & 0.2644\scriptsize ± 0.03\normalsize          & {\underline {0.8627\scriptsize ± 0.02\normalsize}}    & {\underline {0.5972\scriptsize ± 0.04\normalsize}}    & {\underline {0.1916\scriptsize ± 0.02\normalsize}}    \\
\multicolumn{1}{l|}{sVAE+}                                  & \multicolumn{1}{c|}{$3\sigma$}    & 0.0860\scriptsize ± 0.08\normalsize          & -0.0503\scriptsize ± 0.04\normalsize         & 0.0182\scriptsize ± 0.00\normalsize          & 0.3943\scriptsize ± 0.11\normalsize          & 0.0258\scriptsize ± 0.01\normalsize          & 0.0172\scriptsize ± 0.01\normalsize          \\
\multicolumn{1}{l|}{SAMS-VAE}                               &                                   & 0.7603\scriptsize ± 0.03\normalsize          & 0.5299\scriptsize ± 0.04\normalsize          & 0.2828\scriptsize ± 0.03\normalsize          & 0.3381\scriptsize ± 0.32\normalsize          & 0.0509\scriptsize ± 0.05\normalsize          & 0.0546\scriptsize ± 0.04\normalsize          \\
\multicolumn{1}{l|}{\textbf{$\text{\modelname}_{3\sigma}$}} &                                   & \textbf{0.8794\scriptsize ± 0.03\normalsize} & \textbf{0.7292\scriptsize ± 0.06\normalsize} & \textbf{0.3505\scriptsize ± 0.02\normalsize} & \textbf{0.9787\scriptsize ± 0.00\normalsize} & \textbf{0.6469\scriptsize ± 0.05\normalsize} & \textbf{0.4901\scriptsize ± 0.04\normalsize} \\ \hline
\multicolumn{1}{l|}{Conditional VAE}                        &                                   & {\underline {0.7884\scriptsize ± 0.02\normalsize}}    & {\underline {0.5988\scriptsize ± 0.03\normalsize}}    & {\underline {0.3193\scriptsize ± 0.04\normalsize}}    & 0.7119\scriptsize ± 0.07\normalsize          & 0.1434\scriptsize ± 0.02\normalsize          & 0.1005\scriptsize ± 0.03\normalsize          \\
\multicolumn{1}{l|}{CPA-VAE}                                &                                   & 0.7713\scriptsize ± 0.06\normalsize          & 0.5739\scriptsize ± 0.09\normalsize          & 0.2649\scriptsize ± 0.03\normalsize          & {\underline {0.8572\scriptsize ± 0.04\normalsize}}    & {\underline {0.5872\scriptsize ± 0.05\normalsize}}    & {\underline {0.1962\scriptsize ± 0.03\normalsize}}    \\
\multicolumn{1}{l|}{sVAE+}                                  & \multicolumn{1}{c|}{$4\sigma$}    & 0.1020\scriptsize ± 0.09\normalsize          & -0.0513\scriptsize ± 0.05\normalsize         & 0.0192\scriptsize ± 0.00\normalsize          & 0.3866\scriptsize ± 0.10\normalsize          & 0.0266\scriptsize ± 0.01\normalsize          & 0.0186\scriptsize ± 0.01\normalsize          \\
\multicolumn{1}{l|}{SAMS-VAE}                               &                                   & 0.7453\scriptsize ± 0.03\normalsize          & 0.4905\scriptsize ± 0.04\normalsize          & 0.2811\scriptsize ± 0.04\normalsize          & 0.3334\scriptsize ± 0.32\normalsize          & 0.0498\scriptsize ± 0.05\normalsize          & 0.0534\scriptsize ± 0.04\normalsize          \\
\multicolumn{1}{l|}{\textbf{$\text{\modelname}_{4\sigma}$}} &                                   & \textbf{0.8863\scriptsize ± 0.02\normalsize} & \textbf{0.7224\scriptsize ± 0.05\normalsize} & \textbf{0.3686\scriptsize ± 0.02\normalsize} & \textbf{0.9733\scriptsize ± 0.00\normalsize} & \textbf{0.6390\scriptsize ± 0.05\normalsize} & \textbf{0.4866\scriptsize ± 0.05\normalsize} \\ \hline
\multicolumn{1}{l|}{Conditional VAE}                        &                                   & {\underline {0.7991\scriptsize ± 0.02\normalsize}}    & {\underline {0.6143\scriptsize ± 0.03\normalsize}}    & {\underline {0.3302\scriptsize ± 0.03\normalsize}}    & 0.7082\scriptsize ± 0.07\normalsize          & 0.1365\scriptsize ± 0.02\normalsize          & 0.0983\scriptsize ± 0.03\normalsize          \\
\multicolumn{1}{l|}{CPA-VAE}                                &                                   & 0.7813\scriptsize ± 0.05\normalsize          & 0.5915\scriptsize ± 0.08\normalsize          & 0.2793\scriptsize ± 0.03\normalsize          & {\underline {0.8657\scriptsize ± 0.03\normalsize}}    & {\underline {0.5813\scriptsize ± 0.04\normalsize}}    & {\underline {0.1967\scriptsize ± 0.02\normalsize}}    \\
\multicolumn{1}{l|}{sVAE+}                                  & \multicolumn{1}{c|}{$5\sigma$}    & 0.1184\scriptsize ± 0.10\normalsize          & -0.0646\scriptsize ± 0.05\normalsize         & 0.0190\scriptsize ± 0.00\normalsize          & 0.3955\scriptsize ± 0.11\normalsize          & 0.0274\scriptsize ± 0.01\normalsize          & 0.0177\scriptsize ± 0.01\normalsize          \\
\multicolumn{1}{l|}{SAMS-VAE}                               &                                   & 0.7545\scriptsize ± 0.03\normalsize          & 0.4986\scriptsize ± 0.04\normalsize          & 0.2933\scriptsize ± 0.04\normalsize          & 0.3424\scriptsize ± 0.33\normalsize          & 0.0498\scriptsize ± 0.05\normalsize          & 0.0556\scriptsize ± 0.04\normalsize          \\
\multicolumn{1}{l|}{\textbf{$\text{\modelname}_{5\sigma}$}} &                                   & \textbf{0.8716\scriptsize ± 0.03\normalsize} & \textbf{0.7092\scriptsize ± 0.06\normalsize} & \textbf{0.3714\scriptsize ± 0.03\normalsize} & \textbf{0.9635\scriptsize ± 0.01\normalsize} & \textbf{0.4933\scriptsize ± 0.07\normalsize} & \textbf{0.4236\scriptsize ± 0.08\normalsize} \\ \hline

\hline
\multicolumn{2}{c|}{\textbf{Dataset}}                                                           & \multicolumn{3}{c|}{\textbf{Replogle}}                                                                                                     & \multicolumn{3}{c}{\textbf{Adamson}}                                                                                                       \\ \hline
\multicolumn{1}{c|}{Model}                                  & \multicolumn{1}{c|}{QC threshold} & ATE-$\rho$ Top 20                            & ATE-$R^2$ Top 20                             & Jaccard Top 20                               & ATE-$\rho$ Top 20                            & ATE-$R^2$ Top 20                             & Jaccard Top 20                               \\ \hline
\multicolumn{1}{l|}{Conditional VAE}                        &                                   & \textbf{0.8853\scriptsize ± 0.00\normalsize} & \textbf{0.7308\scriptsize ± 0.01\normalsize} & \textbf{0.2939\scriptsize ± 0.00\normalsize} & {\underline {0.8881\scriptsize ± 0.01\normalsize}}    & 0.6639\scriptsize ± 0.02\normalsize          & \textbf{0.3704\scriptsize ± 0.01\normalsize} \\
\multicolumn{1}{l|}{CPA-VAE}                                &                                   & 0.7901\scriptsize ± 0.01\normalsize          & 0.6056\scriptsize ± 0.01\normalsize          & 0.1613\scriptsize ± 0.01\normalsize          & 0.8659\scriptsize ± 0.01\normalsize          & {\underline {0.7039\scriptsize ± 0.02\normalsize}}    & 0.2221\scriptsize ± 0.01\normalsize          \\
\multicolumn{1}{l|}{sVAE+}                                  & \multicolumn{1}{c|}{$3\sigma$}    & 0.7590\scriptsize ± 0.01\normalsize          & 0.4985\scriptsize ± 0.01\normalsize          & 0.1650\scriptsize ± 0.00\normalsize          & 0.8236\scriptsize ± 0.01\normalsize          & 0.6101\scriptsize ± 0.03\normalsize          & 0.1897\scriptsize ± 0.01\normalsize          \\
\multicolumn{1}{l|}{SAMS-VAE}                               &                                   & 0.8291\scriptsize ± 0.01\normalsize          & 0.6364\scriptsize ± 0.02\normalsize          & 0.2600\scriptsize ± 0.02\normalsize          & 0.7349\scriptsize ± 0.02\normalsize          & 0.3117\scriptsize ± 0.01\normalsize          & 0.2203\scriptsize ± 0.01\normalsize          \\
\multicolumn{1}{l|}{\textbf{$\text{\modelname}_{3\sigma}$}} &                                   & {\underline {0.8800\scriptsize ± 0.01\normalsize}}    & {\underline {0.6951\scriptsize ± 0.01\normalsize}}    & {\underline {0.2776\scriptsize ± 0.01\normalsize}}    & \textbf{0.9058\scriptsize ± 0.00\normalsize} & \textbf{0.7860\scriptsize ± 0.01\normalsize} & {\underline {0.3659\scriptsize ± 0.01\normalsize}}    \\ \hline
\multicolumn{1}{l|}{Conditional VAE}                        &                                   & \textbf{0.9032\scriptsize ± 0.00\normalsize} & \textbf{0.7489\scriptsize ± 0.01\normalsize} & \textbf{0.3083\scriptsize ± 0.00\normalsize} & {\underline {0.8869\scriptsize ± 0.01\normalsize}}    & 0.6581\scriptsize ± 0.02\normalsize          & \textbf{0.3712\scriptsize ± 0.01\normalsize} \\
\multicolumn{1}{l|}{CPA-VAE}                                &                                   & 0.8121\scriptsize ± 0.01\normalsize          & 0.6372\scriptsize ± 0.01\normalsize          & 0.1714\scriptsize ± 0.01\normalsize          & 0.8665\scriptsize ± 0.01\normalsize          & {\underline {0.7013\scriptsize ± 0.02\normalsize}}    & 0.2241\scriptsize ± 0.01\normalsize          \\
\multicolumn{1}{l|}{sVAE+}                                  & \multicolumn{1}{c|}{$4\sigma$}    & 0.7893\scriptsize ± 0.01\normalsize          & 0.5333\scriptsize ± 0.01\normalsize          & 0.1771\scriptsize ± 0.00\normalsize          & 0.8224\scriptsize ± 0.02\normalsize          & 0.6103\scriptsize ± 0.03\normalsize          & 0.1946\scriptsize ± 0.01\normalsize          \\
\multicolumn{1}{l|}{SAMS-VAE}                               &                                   & 0.8544\scriptsize ± 0.01\normalsize          & 0.6652\scriptsize ± 0.02\normalsize          & 0.2762\scriptsize ± 0.02\normalsize          & 0.7315\scriptsize ± 0.02\normalsize          & 0.3026\scriptsize ± 0.01\normalsize          & 0.2157\scriptsize ± 0.01\normalsize          \\
\multicolumn{1}{l|}{\textbf{$\text{\modelname}_{4\sigma}$}} &                                   & {\underline {0.8984\scriptsize ± 0.01\normalsize}}    & {\underline {0.7195\scriptsize ± 0.01\normalsize}}    & {\underline {0.3028\scriptsize ± 0.01\normalsize}}    & \textbf{0.9062\scriptsize ± 0.00\normalsize} & \textbf{0.7817\scriptsize ± 0.01\normalsize} & {\underline {0.3516\scriptsize ± 0.01\normalsize}}    \\ \hline
\multicolumn{1}{l|}{Conditional VAE}                        &                                   & \textbf{0.9065\scriptsize ± 0.00\normalsize} & \textbf{0.7454\scriptsize ± 0.01\normalsize} & \textbf{0.3129\scriptsize ± 0.00\normalsize} & {\underline {0.8875\scriptsize ± 0.01\normalsize}}    & 0.6523\scriptsize ± 0.02\normalsize          & \textbf{0.3683\scriptsize ± 0.01\normalsize} \\
\multicolumn{1}{l|}{CPA-VAE}                                &                                   & 0.8180\scriptsize ± 0.01\normalsize          & 0.6430\scriptsize ± 0.01\normalsize          & 0.1751\scriptsize ± 0.01\normalsize          & 0.8715\scriptsize ± 0.01\normalsize          & {\underline {0.7027\scriptsize ± 0.02\normalsize}}    & 0.2234\scriptsize ± 0.01\normalsize          \\
\multicolumn{1}{l|}{sVAE+}                                  & \multicolumn{1}{c|}{$5\sigma$}    & 0.7978\scriptsize ± 0.01\normalsize          & 0.5398\scriptsize ± 0.01\normalsize          & 0.1823\scriptsize ± 0.00\normalsize          & 0.8288\scriptsize ± 0.02\normalsize          & 0.6139\scriptsize ± 0.03\normalsize          & 0.1940\scriptsize ± 0.01\normalsize          \\
\multicolumn{1}{l|}{SAMS-VAE}                               &                                   & 0.8605\scriptsize ± 0.01\normalsize          & 0.6658\scriptsize ± 0.02\normalsize          & 0.2810\scriptsize ± 0.02\normalsize          & 0.7340\scriptsize ± 0.02\normalsize          & 0.2973\scriptsize ± 0.01\normalsize          & 0.2131\scriptsize ± 0.01\normalsize          \\
\multicolumn{1}{l|}{\textbf{$\text{\modelname}_{5\sigma}$}} &                                   & {\underline {0.9006\scriptsize ± 0.01\normalsize}}    & {\underline {0.7261\scriptsize ± 0.01\normalsize}}    & {\underline {0.3077\scriptsize ± 0.01\normalsize}}    & \textbf{0.9048\scriptsize ± 0.00\normalsize} & \textbf{0.7668\scriptsize ± 0.01\normalsize} & {\underline {0.3354\scriptsize ± 0.00\normalsize}}    \\ \hline
\end{tabular}

}
}
\caption{Quantitative evaluation on Norman dataset, Dixit dataset, Replogle dataset, and Adamson dataset across 3$\sigma$, 4$\sigma$, 5$\sigma$ quality control (QC) thresholds. Top 20 results of ATE-$\rho$, ATE-$R^2$, and Jaccard are included in the table. Best results are in bold-faced while second-best ones are underlined.}
\end{table*}

\begin{table}[hbt!]
{\large
\resizebox*{0.75\textwidth}{!}{%
\renewcommand{\arraystretch}{1.15}

\begin{tabular}{l|c|lllll}
\hline
\multicolumn{1}{c|}{Model} & QC threshold & \multicolumn{1}{c}{ATE-$\rho$}               & \multicolumn{1}{c}{ATE-$R^2$}                & \multicolumn{1}{c}{${\text{QCPR}}_{3\sigma}$ (\%)}   & \multicolumn{1}{c}{${\text{QCPR}}_{4\sigma}$ (\%)}   & \multicolumn{1}{c}{${\text{QCPR}}_{5\sigma}$ (\%)}   \\ \hline
                           & $3\sigma$    & 0.4385\scriptsize ± 0.04\normalsize          & 0.1911\scriptsize ± 0.03\normalsize          & \textbf{92.30\scriptsize ± 0.47\normalsize} & \textbf{95.76\scriptsize ± 0.22\normalsize} & \textbf{96.76\scriptsize ± 0.19\normalsize} \\
SAMS-VAE                   & $4\sigma$    & {\underline {0.4573\scriptsize ± 0.05\normalsize}}    & {\underline {0.2047\scriptsize ± 0.04\normalsize}}    & {\underline {89.13\scriptsize ± 0.69\normalsize}}    & {\underline {94.42\scriptsize ± 0.31\normalsize}}    & {\underline {96.18\scriptsize ± 0.16\normalsize}}    \\
                           & $5\sigma$    & \textbf{0.4697\scriptsize ± 0.03\normalsize} & \textbf{0.2125\scriptsize ± 0.03\normalsize} & 87.45\scriptsize ± 0.61\normalsize          & 93.36\scriptsize ± 0.34\normalsize          & 95.49\scriptsize ± 0.21\normalsize          \\ \hline
\end{tabular}
\vspace{1em}
}
}
\vspace{1em}
\caption{Proof of Concept for Quality Control on Norman dataset using SAMS-VAE model.}
\end{table}

\newpage
\section{Qualitative Analysis}
\subsection{Violinplots}
\begin{table*}[hbt!]
\centering
{\Large
\resizebox*{0.55\columnwidth}{!}{%
\renewcommand{\arraystretch}{1.2}
\begin{tabular}{lc}
\hline
\multicolumn{2}{c}{\textbf{Replogle Dataset}}                                                           \\ \hline
\multicolumn{1}{c|}{\textbf{Perturbation}}                                     & \textbf{\# of Samples} \\ \hline
\multicolumn{1}{l|}{Non-targeting}                                             & 2,000                  \\
\multicolumn{1}{l|}{GPS1\_+\_80010011.23-P1P2|GPS1\_-\_80009799.23-P1P2}       & 671                    \\
\multicolumn{1}{l|}{FBXL14\_+\_1703640.23-P1P2|FBXL14\_-\_1703695.23-P1P2}     & 648                    \\
\multicolumn{1}{l|}{NCBP2\_-\_196669400.23-P1P2|NCBP2\_-\_196669410.23-P1P2}   & 547                    \\
\multicolumn{1}{l|}{...}                                                       & ...                    \\
\multicolumn{1}{l|}{UMPS\_-\_124449324.23-P1P2|UMPS\_-\_124449321.23-P1P2}     & 27                     \\
\multicolumn{1}{l|}{PDCD11\_-\_105156445.23-P1P2|PDCD11\_+\_105156417.23-P1P2} & 26                     \\
\multicolumn{1}{l|}{RPS27A\_-\_55459862.23-P1P2|RPS27A\_+\_55459832.23-P1P2}   & 26                     \\
\multicolumn{1}{l|}{PRPF4\_-\_116037989.23-P1P2|PRPF4\_-\_116037979.23-P1P2}   & 25                     \\ \hline
\end{tabular}
}
}
\caption{Summary of Replogle perturbation sample counts. The mean and median are 164 and 144, respectively.}
\end{table*}
\begin{table*}[hbt!]
\centering
{\Large
\resizebox*{0.8\columnwidth}{!}{%
\renewcommand{\arraystretch}{1.2}
\begin{tabular}{lccc}
\hline
\multicolumn{4}{c}{\textbf{Replogle Dataset}}                                                                                                                                   \\ \hline
\multicolumn{1}{c|}{\textbf{Perturbation}}                                       & \textbf{QC Pass rate($\downarrow$)} & \textbf{UMI count} & \textbf{\# of Samples} \\ \hline
\multicolumn{1}{l|}{NPC1\_-\_21166414.23-P1P2|NPC1\_-\_21166384.23-P1P2}         & 0.893204                                       & 5716.663086        & 103                    \\
\multicolumn{1}{l|}{CMTR2\_+\_71323114.23-P1P2|CMTR2\_-\_71323260.23-P1P2}       & 0.864286                                       & 5351.710938        & 140                    \\
\multicolumn{1}{l|}{LAMTOR3\_-\_100815552.23-P1P2|LAMTOR3\_-\_100815661.23-P1P2} & 0.861702                                       & 5587.888672        & 94                     \\
\multicolumn{1}{l|}{RPS18\_+\_33239917.23-P1P2|RPS18\_+\_33239879.23-P1P2}       & 0.853659                                       & 4906.799805        & 41                     \\
\multicolumn{1}{l|}{SYNJ2\_-\_158403158.23-P1P2|SYNJ2\_+\_158402943.23-P1P2}     & 0.850746                                       & 5531.608398        & 201                    \\
\multicolumn{1}{l|}{IPO13\_-\_44412643.23-P1P2|IPO13\_-\_44412666.23-P1P2}       & 0.850000                                       & 5006.784180        & 60                     \\
\multicolumn{1}{l|}{LAMTOR4\_+\_99746556.23-P1P2|LAMTOR4\_-\_99746568.23-P1P2}   & 0.845118                                       & 5434.123535        & 297                    \\
\multicolumn{1}{l|}{SUGP1\_-\_19431214.23-P1P2|SUGP1\_-\_19431183.23-P1P2}       & 0.838951                                       & 5568.571289        & 267                    \\
\multicolumn{1}{l|}{PCNXL3\_+\_65383286.23-P1P2|PCNXL3\_+\_65383293.23-P1P2}     & 0.838384                                       & 5266.072266        & 198                    \\
\multicolumn{1}{l|}{MRPL43\_+\_102747000.23-P1P2|MRPL43\_-\_102747237.23-P1P2}   & 0.838028                                       & 5665.277344        & 142                    \\
\multicolumn{1}{l|}{HIPK3\_-\_33278907.23-P1|HIPK3\_-\_33279054.23-P1}           & 0.836299                                       & 5188.957520        & 281                    \\
\multicolumn{1}{l|}{INO80\_+\_41408265.23-P1P2|INO80\_+\_41408150.23-P1P2}       & 0.835664                                       & 4996.832520        & 286                    \\
\multicolumn{1}{l|}{NFRKB\_+\_129765383.23-P1P2|NFRKB\_+\_129765408.23-P1P2}     & 0.835470                                       & 5163.051270        & 468                    \\
\multicolumn{1}{l|}{CFDP1\_-\_75448185.23-P2|CFDP1\_-\_75448478.23-P2}           & 0.834783                                       & 5603.270996        & 115                    \\
\multicolumn{1}{l|}{TCP1\_-\_160210626.23-P1P2|TCP1\_+\_160210609.23-P1P2}       & 0.833333                                       & 5776.240234        & 30                     \\ \hline
\end{tabular}
}
}
\caption{Summary of QC pass rate, UMI count, and the number of samples used for the Top 15 QC passed Replogle perturbations in our experiments.}
\end{table*}
\begin{table*}[hbt!]
\centering
{\Large
\resizebox*{0.8\columnwidth}{!}{%
\renewcommand{\arraystretch}{1.2}
\begin{tabular}{lccc}
\hline
\multicolumn{4}{c}{\textbf{Replogle Dataset}}                                                                                                                                \\ \hline
\multicolumn{1}{c|}{\textbf{Perturbation}}                                   & \textbf{QC Pass rate ($\uparrow$)} & \textbf{UMI count} & \textbf{\# of Samples} \\ \hline
\multicolumn{1}{l|}{HSPE1\_-\_198365117.23-P1P2|HSPE1\_+\_198365089.23-P1P2} & 0.076923                                        & 2645.000000        & 39                     \\
\multicolumn{1}{l|}{POLD3\_+\_74303696.23-P1P2|POLD3\_-\_74303671.23-P1P2}   & 0.129412                                        & 4710.090820        & 85                     \\
\multicolumn{1}{l|}{DNAJA3\_+\_4475898.23-P1P2|DNAJA3\_-\_4475855.23-P1P2}   & 0.157233                                        & 5338.919922        & 159                    \\
\multicolumn{1}{l|}{POLRMT\_+\_633505.23-P1P2|POLRMT\_+\_633481.23-P1P2}     & 0.188312                                        & 4262.120605        & 308                    \\
\multicolumn{1}{l|}{GINS4\_-\_41386785.23-P1P2|GINS4\_+\_41386860.23-P1P2}   & 0.208955                                        & 5788.071289        & 67                     \\
\multicolumn{1}{l|}{POLD1\_-\_50887659.23-P1P2|POLD1\_-\_50887603.23-P1P2}   & 0.214286                                        & 6473.833496        & 112                    \\
\multicolumn{1}{l|}{MCM2\_-\_127317301.23-P1P2|MCM2\_-\_127317312.23-P1P2}   & 0.225564                                        & 6753.700195        & 133                    \\
\multicolumn{1}{l|}{GAB2\_-\_78128828.23-P1P2|GAB2\_-\_78128897.23-P1P2}     & 0.238208                                        & 4915.900879        & 424                    \\
\multicolumn{1}{l|}{INPPL1\_+\_71935916.23-P1P2|INPPL1\_-\_71935867.23-P1P2} & 0.248555                                        & 4842.813965        & 173                    \\
\multicolumn{1}{l|}{POLR1D\_+\_28196016.23-P1|POLR1D\_+\_28196036.23-P1}     & 0.250000                                        & 4862.000000        & 36                     \\
\multicolumn{1}{l|}{CHAF1A\_-\_4402710.23-P1P2|CHAF1A\_+\_4402728.23-P1P2}   & 0.257143                                        & 7547.111328        & 35                     \\
\multicolumn{1}{l|}{UMPS\_-\_124449324.23-P1P2|UMPS\_-\_124449321.23-P1P2}   & 0.259259                                        & 3764.285645        & 27                     \\
\multicolumn{1}{l|}{MTPAP\_-\_30638029.23-P1P2|MTPAP\_-\_30638037.23-P1P2}   & 0.259740                                        & 4256.299805        & 154                    \\
\multicolumn{1}{l|}{EP400\_+\_132434542.23-P1P2|EP400\_-\_132434629.23-P1P2} & 0.265060                                        & 6353.136230        & 83                     \\
\multicolumn{1}{l|}{LRPPRC\_+\_44223082.23-P1P2|LRPPRC\_-\_44223078.23-P1P2} & 0.267176                                        & 4731.856934        & 131                    \\ \hline
\end{tabular}
}
}
\caption{Summary of QC pass rate, UMI count, and the number of samples used for the Bottom 15 QC Passed (Top 15 QC Failed) Replogle perturbations in our experiments.}
\end{table*}

\newpage
For a better explanation of our model, we selected 4 different gene perturbations by sample counts, QC pass rate, and UMI count as depicted in Table 1-3. Specifically, we chose non-targeting with the highest sample count 2000, GAB2 having low QC pass rate (23.82\%) and high sample counts (424), NFRKB with high QC pass rate (83.55\%) and sample counts (468), PRPF4 with low sample counts (25), and showed violin plots for each QC sub-criteria as shown in our main Figure 3 for all 4 cases.

\begin{figure*}[!t]
\centering
\begin{minipage}[b]{\textwidth}
    \centering
    \includegraphics[width=\textwidth]{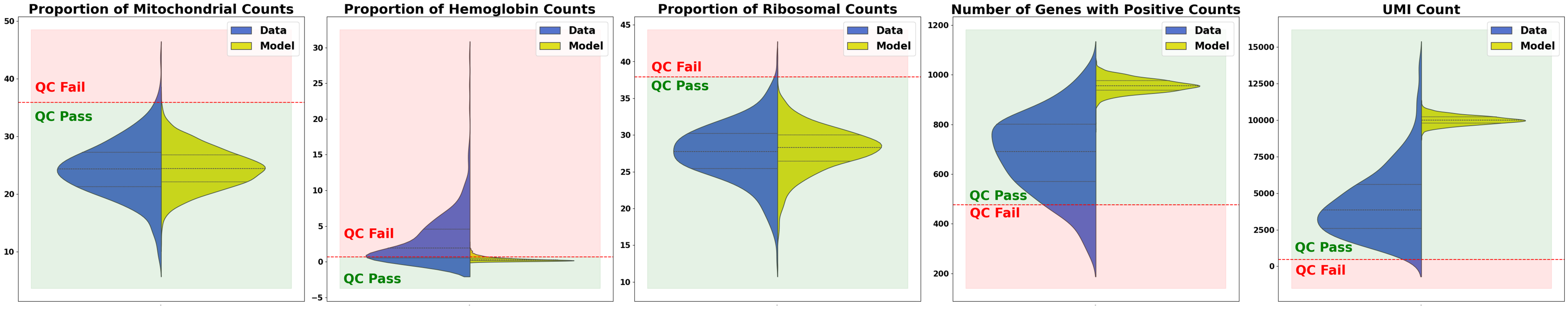}
    \caption{Violin plots of GAB2-perturbed cellular response for each QC sub-criteria.}   
\end{minipage}
\vspace{1em}
\begin{minipage}[b]{\textwidth}
    \centering
    \includegraphics[width=\textwidth]{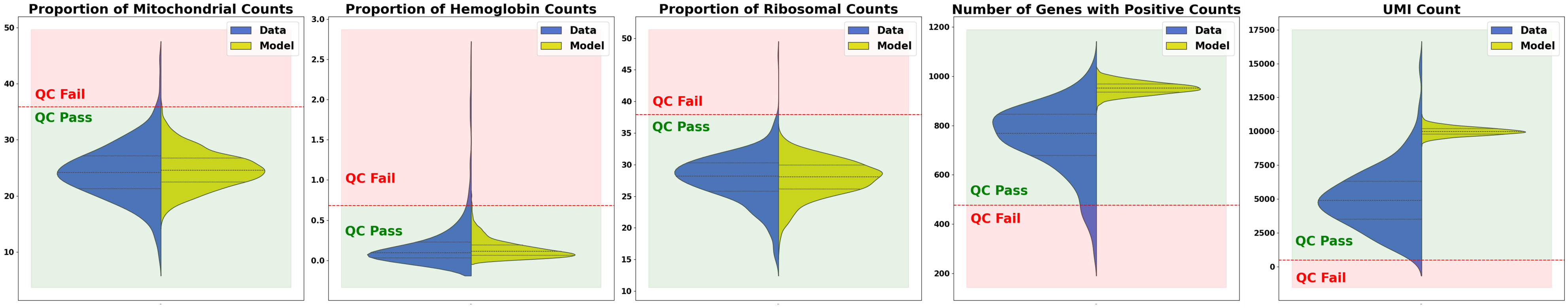}
    \caption{Violin plots of NFRKB-perturbed cellular response for each QC sub-criteria.}   
\end{minipage}
\vspace{1em}
\begin{minipage}[b]{\textwidth}
    \centering
    \includegraphics[width=\textwidth]{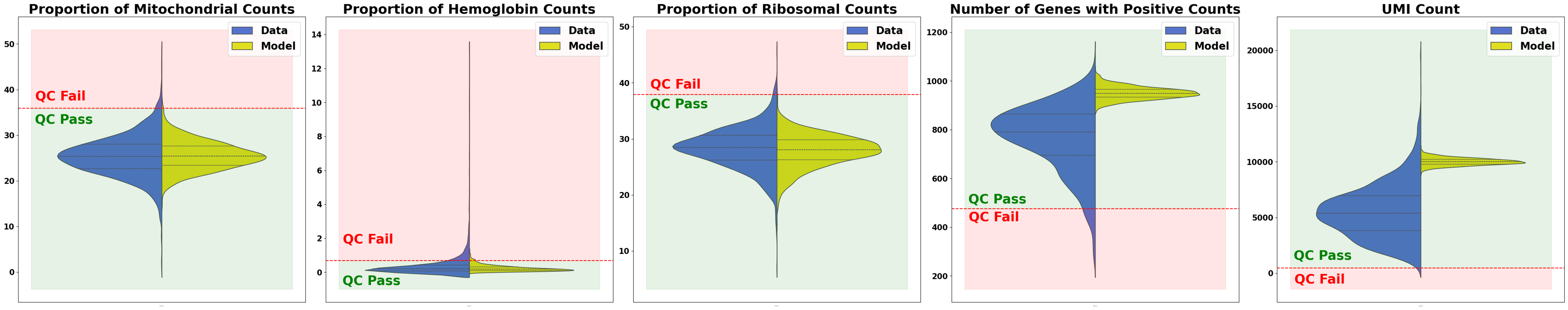}
    \caption{Violin plots of Non-targeting control cellular response for each QC sub-criteria.}   
\end{minipage}
\vspace{1em}
\begin{minipage}[b]{\textwidth}
    \centering
    \includegraphics[width=\textwidth]{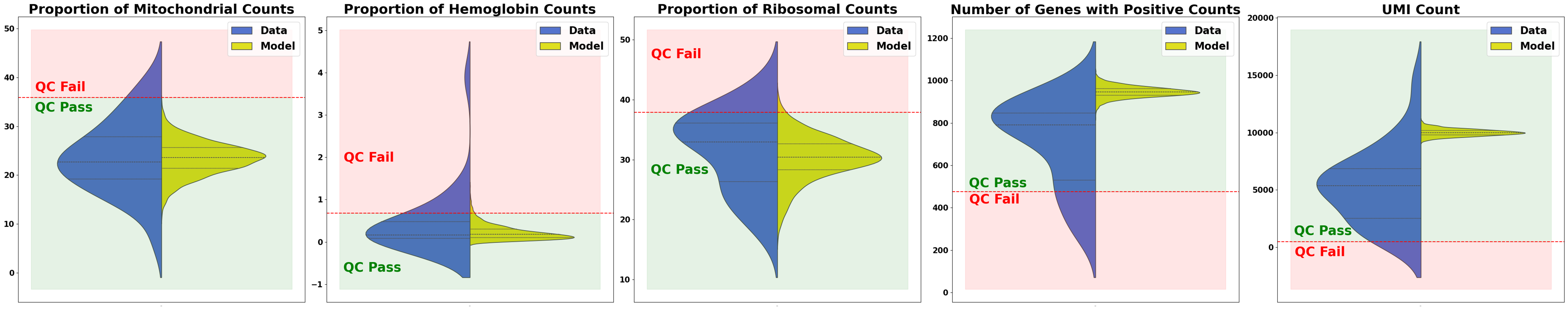}
    \caption{Violin plots of PRPF4-perturbed cellular response for each QC sub-criteria.}   
\end{minipage}
\vspace{1em}

\end{figure*}

\newpage
\subsection{UMAP of latent basal state embeddings}
As depicted in Figure 5, the UMAP of latent basal state embeddings is not distinguished between artifacts and perturbations. This suggests that both perturbations and artifacts have been effectively disentangled from the basal state embeddings, resulting in non-distinguishable basal state embeddings.

\begin{figure}
\centering
\includegraphics[width=\textwidth]{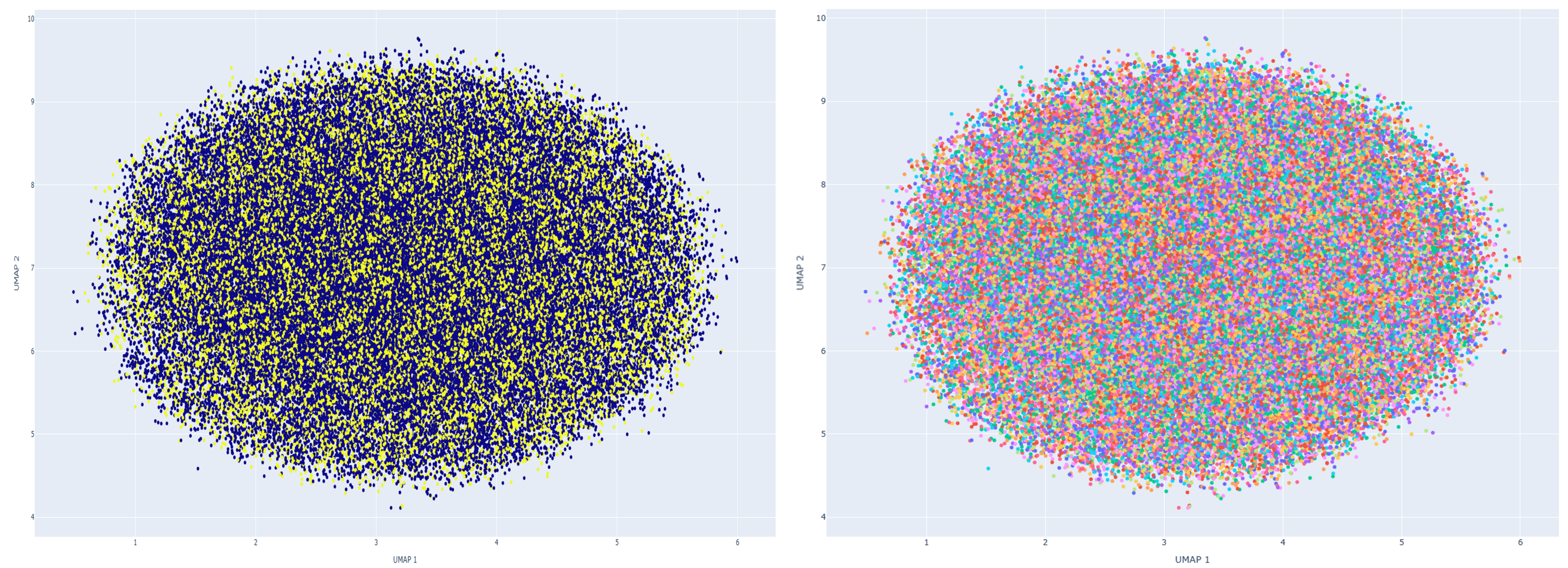}
\caption{UMAP of latent basal state embeddings $z^b$ labeled by artifact presence (left) and perturbation type (right).}
\end{figure}

\newpage
\section{Experiment Details}
\subsection{Norman}
Each model was optimized with the Adam optimizer for 2,000 epochs with a batch size of 512, learning rate of 0.0003, and gradient clipping norm of 100. The data was processed using the NormanOODCombinationDataModule, with 75\% of the data allocated for training and 25\% for testing. We varied the split seed across {0, 1, 2, 3, 4} to evaluate robustness. Additionally, we considered quality control (QC) thresholds of {3, 4, 5}, training the model separately for each threshold and evaluating them all individually. For the model, we used the \textsc{CradleVAE} Model with a latent dimension of 200 and one decoder layer. The prior probability of the mask was set to 0.01, and the embedding prior scale to 1. The guide utilized was \textsc{CradleVAE} CorrelatedNormalGuide, with 4 layers and 400 hidden units in the embedding encoder, and the basal encoder input was normalized using log standardize. The loss function was defined by \textsc{CradleVAE} ELBOLossModule with $\beta$ = 0.5. We observed that these settings provided a balanced performance across the experiments. In addition, we employed the \textsc{CradleVAE} Predictor for evaluation purposes.

\subsection{Dixit}
Each model was optimized with the Adam optimizer for 2,000 epochs, using a batch size of 512, a learning rate of 0.0003, and a gradient clipping norm of 100. Data was processed using the DixitOODCombinationDataModule, with 75\% for training and 25\% for testing. We varied the split seed across {0, 1, 2, 3, 4} and considered QC thresholds of {3, 4, 5}, training and evaluating the model separately for each threshold. The model used was \textsc{CradleVAE} Model with a latent dimension of 200, one decoder layer, a mask prior probability of 0.01, and an embedding prior scale of 1. The guide was \textsc{CradleVAE} CorrelatedNormalGuide with 4 layers and 400 hidden units, using log standardize for input normalization. The loss function was \textsc{CradleVAE} ELBOLossModule with $\beta$ = 0.5, and the lightning module had a learning rate of 0.0003 and 5 particles.

\subsection{Replogle}
Each model was optimized with the Adam optimizer for 2,000 epochs, using a batch size of 512, a learning rate of 0.001, and a gradient clipping norm of 100. Data was processed using the ReplogleDataModule, and we varied the split seed across {0, 1, 2, 3, 4}. QC thresholds of {3, 4, 5} were considered, training and evaluating the model separately for each. The model used was \textsc{CradleVAE} Model with a latent dimension of 200, one decoder layer, a mask prior probability of 0.001, and an embedding prior scale of 1. The guide was \textsc{CradleVAE} CorrelatedNormalGuide, featuring 4 layers and 400 hidden units, with input normalization using log standardize. The loss function was \textsc{CradleVAE} ELBOLossModule with $\beta$ = 0.05.

\subsection{Adamson}
Each model was optimized with the Adam optimizer for 2,000 epochs, using a batch size of 512, a learning rate of 0.0001, and a gradient clipping norm of 100. Data was processed using the AdamsonDataModule, with varying split seeds {0, 1, 2, 3, 4}. QC thresholds of {3, 4, 5} were evaluated, with separate training and assessment for each threshold. The model employed was \textsc{CradleVAE} Model with a latent dimension of 100, one decoder layer, a mask prior probability of 0.001, and an embedding prior scale of 1. The guide used was \textsc{CradleVAE} CorrelatedNormalGuide, featuring 4 layers and 400 hidden units, with input normalization using log standardize. The loss function was \textsc{CradleVAE} ELBOLossModule with $\beta$ = 0.1.

\subsection{Implementation Details}
For the baselines, the parameters were configured as specified in the original papers. All experiments were conducted on a Ubuntu server with a single NVIDIA RTX 3090Ti GPU and 24 GB memory size.

\end{appendices}
\newpage

\end{document}